\documentclass[letterpaper, paper,11pt]{AAS}

\usepackage{amsmath}
\usepackage{amssymb}
\usepackage{mathtools}
\usepackage{bm}
\newcommand {\R}{\mathbb{R}}
\DeclareMathOperator*{\argmax}{arg\,max}

\usepackage{algorithm}
\usepackage{algorithmic}

\makeatletter
\newcommand\fs@norules{\def\@fs@cfont{\bfseries}\let\@fs@capt\floatc@ruled
  \def\@fs@pre{}%
  \def\@fs@post{}%
  \def\@fs@mid{\kern3pt}%
  \let\@fs@iftopcapt\iftrue}
\makeatother
\restylefloat{algorithm}

\usepackage{caption}
\usepackage{subcaption}

\usepackage{url}

\newcommand{\TE}{\mathrm{TE}}
\newcommand{\CT}{\mathrm{CT}}
\newcommand{\CE}{\mathrm{CE}}
\newcommand{\OC}{\mathrm{OC}}
\newcommand{\OT}{\mathrm{OT}}
\newcommand{\C}{\mathrm{C}}

\usepackage[colorlinks=true, pdfstartview=FitV, linkcolor=black, citecolor= black, urlcolor= black]{hyperref}
\usepackage{overcite}
\usepackage{footnpag}			      	

\PaperNumber{24-200}

\begin{document}

\title{Factor Graph-Based Active SLAM for Spacecraft Proximity Operations}

\author{Lorenzo Ticozzi\thanks{PhD candidate, School
of Aerospace Engineering, Georgia Institute of Technology, Atlanta,
GA, 30332-0150, USA. lorenzo@gatech.edu},  
Panagiotis Tsiotras\thanks{David and Andrew Lewis Chair and Professor, School
of Aerospace Engineering, Georgia Institute of Technology, Atlanta,
GA, 30332-0150, USA. tsiotras@gatech.edu}
}

\maketitle{}

\begin{abstract}
We investigate a scenario where a chaser spacecraft or satellite equipped with a monocular camera navigates in close proximity to a target spacecraft. 
The satellite's primary objective is to construct a representation of the operational environment and localize itself within it, utilizing the available image data.
We frame the joint task of state trajectory and map estimation as an instance of smoothing-based simultaneous localization and mapping (SLAM), where the underlying structure of the problem is represented as a factor graph.
Rather than considering estimation and planning as separate tasks, we propose to control the camera observations to actively reduce the uncertainty of the estimation variables, the spacecraft state, and the map landmarks.
This is accomplished by adopting an information-theoretic metric to reason about the impact of candidate actions on the evolution of the belief state. 
Numerical simulations indicate that the proposed method successfully captures the interplay between planning and estimation, hence yielding reduced uncertainty and higher accuracy when compared to commonly adopted passive sensing strategies.
\end{abstract}

\section{Introduction}
In the last decade, the space sector has been characterized by a resurgence of interest from both public and private stakeholders. 
On one hand, the growing demand for space-based services has led to a continuous growth in the number of operational satellites orbiting the Earth for commercial purposes \cite{Wilde2019editorial}; on the other, governmental organizations worldwide are setting their priorities towards an enhanced exploitation of the extraterrestrial segment, whether for scientific or military purposes.
As a result, the development of in-space servicing, assembly and manufacturing (ISAM) capabilities has emerged as a critical goal \cite{NSTC2022isam} for strategic, economic and environmental reasons alike.
ISAM operations encompass a spectrum of activities, ranging from satellite inspection, refuelling and repair, to the assembly and maintenance of large orbiting structures such as the proposed lunar Gateway \cite{Creech2022Artemis}.
A fundamental requirement for these operations is the ability to perform accurate autonomous relative navigation between spacecraft — referred to as ``chasers'' or ``servicers'' — and resident space objects (RSOs) of interest.

\subsection{Prior Work}
The literature on relative navigation techniques is diverse, primarily distinguished by the types of sensors and estimation methods used. 
Traditional approaches often combine filters with various sensor configurations. For instance, Aghili introduced a robust 3D pose estimation method that synergizes a range laser camera with an extended Kalman filter (EKF). \cite{Aghili2011FaultTolerant}
This combination accelerates convergence by using the filter's output as an initial guess for the Iterative Closest Point (ICP) algorithm, which performs point cloud registration. 
The enhanced pose from the ICP, in turn, feeds back into the filter.
Addressing the challenges of ICP initialization and computational load, Opromolla et al.~propose a template-matching strategy. \cite{Opromolla2015UncoopPose}
This method leverages prior knowledge of the target's 3D shape for initial pose estimation, offering a more efficient approach.
Vision-based systems present a less computationally demanding alternative, capable of long-range measurements, thus being well-suited for spacecraft proximity operations. Their effectiveness is evidenced in missions like ETS-VII, which utilized the Proximity Camera Sensor \cite{Mokuno2011OpticalNav}, or the Orbital Express mission in 2007,  which was facilitated by the Advanced Video Guidance Sensor \cite{Friend2008OE}.

The GNC community has shown considerable interest in both monocular and stereo-vision solutions. 
Fasano et al.~detail a stereo vision system designed for close-proximity cooperative scenarios, integrating stereo measurements into the correction stage of a Kalman filter and employing Clohessy-Wiltshire equations for prediction. \cite{Fasano2009Infotech} 
Segal et al.~extend this approach, using multiple iterated EKFs to estimate the inertia tensor of the target. \cite{Segal2014Stereovision}
Building on these concepts, Pesce et al.~introduce a pseudo-measurement in the iterated EKF, simplifying the estimation of the target's inertia ratios. \cite{Pesce2017Stereovision}

Recently, two solutions have gained popularity as alternatives to traditional navigation filters. First, machine learning (ML) techniques aim to learn mappings from images to pose transformations, as explored in works like Sharma et al.~and Black et al. \cite{Sharma2020NNPoseEst, Black2021realtime}
However, ML-based methods often require extensive datasets for training, and their performance across different domains (the ``domain gap'') remains a challenge. On the other hand, smoothing techniques offer a simpler structure, adaptable to various scenarios and sensor configurations, often achieving better accuracy than filtering \cite{Strasdat2010SLAM}.

In mobile robotics, Simultaneous Localization and Mapping (SLAM) is often approached as a smoothing problem over measurements and control inputs to infer both the robot state trajectory and an environmental map \cite{Randall1986SLAM, Thrun2003Mapping}. 
Unlike filters, smoothers provide a globally consistent probabilistic representation of the estimation unknowns. 
The probability density over the pose and map variables can be decomposed into conditional densities over smaller variable subsets, a factorization that can be encoded in a graph model \cite{Dellaert2017factor}. 
This operation offers insights into the problem's structure, making graph-based SLAM an ideal framework for describing and solving monocular relative navigation in space.

Tweddle et al.~investigated SLAM in spacecraft proximity operations using a factor graph model that integrates stereo measurements with a probabilistic representation of the target rigid-body dynamics, assuming torque-free motion. \cite{Tweddle2015FGs} 
This approach was validated on the International Space Station (ISS) during the SPHERES experiment \cite{Mohan2009Spheres}. 
Factor graph-based representations also prove effective for navigation near small celestial bodies, as demonstrated by Rathinam et al. \cite{Rathinam2017IAC} and Dor et al. \cite{dor2022astroslam}, by incorporating motion factors from orbital mechanics and visual monocular measurements of asteroid surfaces. 
Furthermore, graph formulations are suitable for distributed navigation scenarios, where periodic graph merging allows agents to share information, as shown by Elankumaran et al. \cite{Elankumaran2021Distributed}.

In all the aforementioned works, SLAM has been formulated in a ``passive'' manner, that is, the agent does not have the ability to affect the decision-making process to better characterize its state and surroundings.
Active SLAM, an extension of passive SLAM, integrates how robotic action impacts the estimation processes, enhancing the agent's awareness through strategic planning. \cite{Placed2023survey}
While active SLAM has received significant attention in mobile ground robotics, only a handful of studies investigate its potential for on-orbit operations.
Among these, some early works follow a similar approach based on Cross Entropy (CE) minimization, parameterizing the control input and then iteratively solving the resulting finite-dimensional optimization as the estimation of a rare event probability. \cite{Kontitsis2016Info, antonello2016performance, antonello2016vision}
However, the necessary evaluations of the uncertainty-aware cost function are predicated on simulating an EKF for each of the sampled trajectories, a procedure that scales poorly as the state dimension increases and suffers from strong nonlinearities.
A sampling-based guidance strategy was proposed \cite{Maestrini2022Guidance} to obtain perception-aware spacecraft proximity trajectories; in particular, a reward function is defined such that the chaser spacecraft is encouraged to visit regions with lower landmark density.
While the approach results in a uniform exploration of the target region, it remains unclear how the proposed strategy would impact the accuracy of the navigation step.
The work by Nakath et al.~is perhaps the most complete contribution to active SLAM in space up-to-date. \cite{Nakath2019Active} 
Herein, the optimal spacecraft trajectory maximizes an expected reward consisting of two terms encoding information gain (exploration) and localization (exploitation).
The solution relies on a high-dimensional grid-like discretization of the space to be mapped, where each grid cell is associated with a belief function modeling the uncertainty about its estimated occupancy.
The information and localization terms for a candidate trajectory are computed by accumulating the non-specificity and normalized occupancy values, respectively, of the grid cells visible along the trajectory.
By assigning different weights to the two gains, the authors effectively drive SLAM towards exploration over exploitation, and vice versa.

In conclusion, previous approaches either rely on an expensive update of the state and/or map uncertainty \cite{Kontitsis2016Info, antonello2016performance, antonello2016vision} or, alternatively, they follow a more indirect approach by encouraging observations of ``promising'' regions of the environment \cite{Maestrini2022Guidance, Nakath2019Active}, possibly renouncing to a more principled approach.

\subsection{Contributions}

In this work, we build upon the success and computational efficiency of smoothing-based visual SLAM formulations and on previous results in terms of belief space planning (BSP) \cite{Kopitkov2017bsp} to propose a factor graph-based active SLAM solution for spacecraft proximity operations.
In particular, we seek spacecraft camera orientations that lead to a maximum reduction in the estimated uncertainty of the joint state and map estimate.
Diverging from previous methodologies \cite{Kontitsis2016Info, antonello2016performance, antonello2016vision, Nakath2019Active}, our approach uniquely computes the information-theoretic reward directly from the structure of the estimation graph. 
This strategy enables the planner to more effectively consider the evolution of the SLAM belief state. 
As a result, we achieve a more integrated and principled connection between the planning and estimation phases, enhancing the overall SLAM performance.
To the best of our knowledge, this is the first work leveraging information-theoretic BSP in the context of spacecraft proximity guidance.

We evaluate the ``informativeness'' of the trajectory candidates by reasoning about the expected \textit{a posteriori} distribution of the estimate, referred to as the belief, which is represented by a factor graph.
By leveraging the standard assumption of Gaussian distribution of the belief, we are able to compute the expected information gain brought 
about by a potential action, which is used to assign scores to a set of sampled trajectories.
A significant advantage of our approach is that the computation of the a posteriori information gain can be made independent from the dimension of the state \cite{Kopitkov2017bsp}.

We validate our approach 
in a close-proximity spacecraft navigation scenario where the planning module onboard the chaser is tasked with obtaining an informative camera-pointing strategy. 
The solutions resulting from our policy are compared to the ones obtained when following two passive camera orientation profiles.
The results support our claim that the proposed method leads to a significant uncertainty reduction in both the state and map variables while also achieving better estimation accuracy.

\section{Problem Statement} 
\label{sec:problem_statement}
In this section, we summarize some necessary elements of relative orbital dynamics along with the relevant mathematical notation to be used in this paper.
Then, we provide a brief overview of factor graphs for SLAM applications.
Finally, we formulate the active SLAM problem for a spacecraft proximity navigation scenario.

\subsection{Preliminaries and Dynamic Model}
\label{sec:prelim_dynamics}
We consider two different spacecraft, a chaser $C$ equipped with a monocular camera and a target $T$.
Both $C$ and $T$ are subject to the gravitational pull of a main attractor body $E$.
We assume a fixed and known transformation between the chaser spacecraft and the camera frame, henceforth referring to them interchangeably.
An inertial reference frame $\mathcal{E}=\{\mathbf{e}_i\}_{i=1}^3$ is fixed in $E$, where each $\mathbf{e}_i \in \R^3$ represents an element of a triad of unit orthogonal vectors.
The orbit of $T$ around $E$ is assumed nearly circular, and its instantaneous position and velocity with respect to $E$ are described by $\mathbf{r}_{\TE}^{\mathcal{E}}=[r_{\TE,x}^{\mathcal{E}}, r_{{\TE},y}^{\mathcal{E}}, r_{\TE,z}^{\mathcal{E}}]^\top$ and $\mathbf{v}_{\TE}^{\mathcal{E}}=[v_{\TE,x}^{\mathcal{E}}, v_{\TE,y}^{\mathcal{E}}, v_{\TE,z}^{\mathcal{E}}]^\top$, respectively, where $\mathbf{r}_{\TE}^{\mathcal{E}}$ is read as ``the position of $T$ w.r.t.~$E$, expressed in frame $\mathcal{E}$'', and similarly for the velocity $\mathbf{v}_{\TE}^{\mathcal{E}}$.
The position of the chaser $C$ is defined in a similar manner as $\mathbf{r}_{\CE}^{\mathcal{E}}$.
A second reference frame centered in $T$ is established, $\mathcal{T}=\{\mathbf{t}_i\}_{i=1}^3$, where
\begin{align}
    \mathbf{t}_1 = \frac{\mathbf{r}_{\TE}^{\mathcal{E}}}{\lVert \mathbf{r}_{\TE}^{\mathcal{E}} \rVert}, \quad
    \mathbf{t}_3 = \frac{
    \mathbf{h}_{\TE}^{\mathcal{E}}
    }{
    \lVert
    \mathbf{h}_{\TE}^{\mathcal{E}}
    \rVert
    }, \quad
    \mathbf{t}_2 = \mathbf{t}_3 \times \mathbf{t}_1,
\end{align}
and $\mathbf{h}_{\TE}^{\mathcal{E}} = \mathbf{r}_{\TE}^{\mathcal{E}} \times \mathbf{v}_{\TE}^{\mathcal{E}}$ is the angular momentum of $T$.
If $\mathbf{r}_{\CT}^{\mathcal{E}} = \mathbf{r}_{\CE}^{\mathcal{E}}-\mathbf{r}_{\TE}^{\mathcal{E}}$ is such that 
$\lVert \mathbf{r}_{\CT}^{\mathcal{E}} \rVert \ll \lVert \mathbf{r}_{\TE}^{\mathcal{E}} \rVert$, then the relative chaser-target dynamics in the $\mathcal{T}$-frame can be described by a system of ODEs known as Clohessy-Wiltshire (CW) equations \cite{Curtis2014OM}, as follows:
\begin{align}
    \dot{v}_{\CT,x}^{\mathcal{T}} &= 3 \nu^2 r_{\CT,x}^{\mathcal{T}} + 2 \nu v_{\CT,y}^{\mathcal{T}} + w_{\C,x}^{\mathcal{T}}, \label{eq:CWx} \\
    \dot{v}_{\CT,y}^{\mathcal{T}} &= -2 \nu v_{\CT,x}^{\mathcal{T}} + w_{\C,y}^{\mathcal{T}}, \label{eq:CWy} \\
    \dot{v}_{\CT,z}^{\mathcal{T}} &= -\nu^2 r_{\CT,z}^{\mathcal{T}} + w_{\C,z}^{\mathcal{T}}, \label{eq:CWz}
\end{align}
where $\nu$ is the orbital angular velocity of $T$ and $\mathbf{w}_{\C}^{\mathcal{T}} = [w_{\C,x}^{\mathcal{T}}, w_{\C,y}^{\mathcal{T}}, w_{\C,z}^{\mathcal{T}}]^{\top}$ is a disturbance acceleration distributed according to a zero-mean Gaussian, $\mathbf{w}_{\C}^{\mathcal{T}} \sim \mathcal{N}(0_{3 \times 1}, \Sigma_{\text{W}})$, with covariance $\Sigma_{\text{W}}$.

In the undisturbed case $\mathbf{w}_{C}^{\mathcal{T}} = 0_{3 \times 1}$, the dynamic equations \eqref{eq:CWx}-\eqref{eq:CWz} 
have a closed-form solution \cite{Curtis2014OM}, which can be used to predict the motion of $C$ relative to $T$.
Moreover, if the initial conditions are such that $v_{\CT,y}^{\mathcal{T}}(t_0) = -2\nu r_{\CT,x}^{\mathcal{T}}(t_0)$, then the relative path of the chaser relative to $T$ is a closed ellipse. 
This condition is of great use since it allows the chaser to move in the proximity of the target for a (potentially) indefinite amount of time without applying significant control effort.

A third reference frame $\mathcal{C}= \{\mathbf{c}_i\}_{i=1}^3$ centered on the chaser spacecraft can be defined as follows
\begin{align} \label{eq:c123}
    \mathbf{c}_3 = \frac{\mathbf{r}_{\OC}^{\mathcal{T}}}{\lVert \mathbf{r}_{\OC}^{\mathcal{T}} \rVert},  \quad
    \mathbf{c}_2 = \frac{\mathbf{v}_{\CT}^{\mathcal{T}} \times \mathbf{r}_{\OC}^{\mathcal{T}}}{
    \lVert \mathbf{v}_{\CT}^{\mathcal{T}} \times \mathbf{r}_{\OC}^{\mathcal{T}} \rVert
    }, \quad
    \mathbf{c}_1 = \mathbf{c}_2 \times \mathbf{c}_3, 
\end{align}
where $\mathbf{r}_{\OC}^{\mathcal{T}}=\mathbf{r}_{\OT}^{\mathcal{T}} - \mathbf{r}_{\CT}^{\mathcal{T}}$, and $\mathbf{r}_{\OT}^{\mathcal{T}}$ describes the position of point $O$, the chaser's camera observation target.
The camera is fixed on the body of the chaser, and its boresight is aligned with $\mathbf{c}_3$.
Therefore, considering $\mathbf{r}_{\CT}^{\mathcal{T}}$ and $\mathbf{v}_{\CT}^{\mathcal{T}}$ as given by the solution to the relative dynamics in \eqref{eq:CWx}-\eqref{eq:CWz}, $\mathcal{C}$ can be fully defined by assigning an observation target $O$ to the camera.

Finally, the full relative pose of the chaser (camera) with respect to the target is represented by $T_{\mathcal{TC}} \in \mathrm{SE}(3)$,
\begin{equation}
\label{eq:sc_state}
    T_{\mathcal{TC}} =
    \begin{bmatrix}
        R_{\mathcal{TC}} & \mathbf{r}_{\CT}^{\mathcal{T}} \\
        0_{1\times 3}   & 1
    \end{bmatrix},
\end{equation}
with $R_{\mathcal{TC}}=\left[\mathbf{c}_1, \mathbf{c}_2, \mathbf{c}_3\right] \in \mathrm{SO}(3)$ the camera attitude.

\subsection{Bayesian Inference and Factor Graphs}
\label{sec:intro_inference}

In this section, we lay the foundations of our formulation of the problem by recalling some ideas at the intersection between SLAM, probabilistic inference, and factor graphs. \cite{Dellaert2017factor}

SLAM can be framed as an instance of probabilistic inference, i.e., the problem of estimating an unknown quantity given a set of observations that lead to some belief about the unknown quantity.
Within the traditional SLAM framework, the unknown variable $X_k = \{T_{0:k}, M_{k}\}$ at time $t_k$, $k=0,\ldots,L$, is given by the robot pose trajectory $T_{0:k}$ and by $M_{k}$, the map landmarks triangulated up to $t_k$.
Given new observations, the belief can be updated, resorting to the well-known Bayes' rule.
For a history of measurements $Z_{0:k}$, the Bayes' rule states that
\begin{equation}
\label{eq:Bayes}
    p(X_k | Z_{0:k}) = \frac{p(Z_{0:k} | X_k) p(X_k)}{p(Z_{0:k})}.
\end{equation}
In fact, since the measurements are given, $p(Z_{0:k})$ only acts as a constant scaling factor on the value of the posterior $p(X_k | Z_{0:k})$. 
Moreover, the conditional $p(Z_{0:k} | X_k)$ can be substituted by the likelihood $l(X_k; Z_{0:k}) \propto p(Z_{0:k} | X_k)$ to highlight the dependence on $X_k$, the unknown of the problem.
Hence, Eq.~\eqref{eq:Bayes} is rewritten as follows,
\begin{equation}
\label{eq:Bayes1}
    p(X_k | Z_{0:k}) \propto l(X_k; Z_{0:k}) p(X_k).
\end{equation}
The estimate of $X_k$ can be inferred through a maximum a posteriori (MAP) estimator, i.e., the maximizer of the a posteriori density $p(X_k | Z_{0:k})$,
\begin{equation}
    X^{\text{MAP}}_k = \argmax_{X_k}  p(X_k | Z_{0:k})  \label{eq:map1} 
    = \argmax_{X_k} l(X_k; Z_{0:k}) p(X_k).
\end{equation}

\paragraph{Factor Graphs.}
Factor graphs are bipartite graphs defined as $\mathcal{F} = (\mathcal{U}, \mathcal{X}, \mathcal{E})$. \cite{Koller2009ProbGraph}
Two types of nodes exist in a factor graph, factor nodes $\phi_i \in \mathcal{U}$ and variable nodes $x_j \in \mathcal{X}$; the edge $\epsilon_{ij} \in \mathcal{E}$ connects the nodes $\phi_i$ and $x_j$.
Factor graphs allow to represent any factored function $\psi(X)$ over a set of variables $X$, as follows
\begin{equation}
    \psi(X) = \prod_{i}\phi_i(X_i),
\end{equation}
where $X_i \subseteq X$ is the subset of variables adjacent to $\phi_i$ in $\mathcal{F}$.
Every Bayes net can be converted to a factor graph where $x_j$ represent estimation variables, while the factors $\phi_i$ represent probabilistic constraints between the variables.

As a consequence, the SLAM formulation in Eqs.~\eqref{eq:Bayes}-\eqref{eq:Bayes1} can be expressed in a different form, which more clearly highlights the dependence relationships between variables and observations.
At time $t_k$, the probability distribution function (pdf) over $X_k$ is factored as follows,
\begin{equation}
\label{eq:factor_graph_pdf}
    \psi(X_k) = \prod_{i=0}^{k} \prod_{j=1}^{n_i} \phi_i^j(X_{i}^j) \propto p(X_k | Z_{0:k}),
\end{equation}
where $\phi_i^j(X_{i}^j)$ is one of the $n_i$ factors added at time $t_i$, which is adjacent to variables $X_i^j$.
Note that, since the observation history $Z_{0:k}$ is given, it does not appear explicitly as a variable in $\psi(X_k)$, but only as a parameter.
In most SLAM contexts, the factors $\phi_i^j(X_{i}^j)$ encode a probabilistic constraint arising from prior knowledge of the problem or from measurements among subsequent robot states (e.g., odometry factors), between a state and a landmark (e.g., camera projection factors), etc.
Moreover, factors are commonly modeled as zero-mean Gaussians,
\begin{equation}
    \label{eq:general_factor}
    \phi_i^j(X_{i}^j) \propto \exp{\left(-\frac{1}{2} \lVert h_i^j(X_i^j) - z_i^j \rVert_{\Sigma_i^j}^2 \right)},
\end{equation}
where, for example, $h_i^j(X_i^j)$ is a nonlinear measurement model and $z_i^j$ is the $j$-th noise-corrupted measurement acquired at $t_i$, that is,
\begin{equation}
    z_i^j = h_i^j(X_i^j) + v_i^j, \quad v_i^j \sim \mathcal{N}(0, \Sigma_i^j).
\end{equation}
Rewriting Eq.~\eqref{eq:map1} using Eqs.~\eqref{eq:factor_graph_pdf}-\eqref{eq:general_factor} yields a MAP inference problem predicated on a factor graph model, as follows
\begin{equation}
    \label{eq:factor_graph_MAP_1}
    X_k^{\text{MAP}} = \argmax_{X_k} \prod_{i=0}^{k} \prod_{j=1}^{n_i} \phi_i^j(X_{i}^j)
    = \argmax_{X_k} \prod_{i=0}^{k} \prod_{j=1}^{n_i} \exp{\left(-\frac{1}{2} \lVert h_i^j(X_i^j) - z_i^j \rVert_{\Sigma_i^j}^2 \right)}.
\end{equation}
Upon linearizing and taking the negative log of the exponential terms, Eq.~\eqref{eq:factor_graph_MAP_1} is transformed into a standard least-squares problem which can be tackled using one of the available software tools for factor graph-based inference, such as the \texttt{GTSAM} library \cite{Dellaert2017factor}.

\subsection{Problem Definition}
\label{sec:problem_def}

In the following, we build upon the notation and ideas introduced in the previous sections to formalize our factor graph-based active SLAM problem for a spacecraft proximity navigation scenario.

At $t_k$, the (true) pose of the chaser is described by $T_{\mathcal{TC}}^k$, whose definition follows from Eq.~\eqref{eq:sc_state}; from now on, we drop the subscripts for brevity and refer to the instantaneous pose of $C$ as $T_k \triangleq T_{\mathcal{TC}}^k$, and, similarly, for $R_k \triangleq R_{\mathcal{TC}}^k$, $\mathbf{r}_k \triangleq \mathbf{r}_{CT}^{\mathcal{T}}(t_k)$, $\mathbf{v}_k \triangleq \mathbf{v}_{CT}^{\mathcal{T}}(t_k)$. 
The chaser is equipped with a camera whose boresight is parallel to the $\mathbf{c}_3$ axis defined in Eq.~\eqref{eq:c123}, as mentioned earlier.

Throughout $T_{0:k}$, a sequence of noisy camera observations $Z_{0:k} = \{Z_0, \ldots, Z_k\}$ is processed, where
\begin{equation}
\label{eq:Z_i}
    Z_i = \bigcup_{j=1}^{n_i} z_i^j
    = \bigcup_{j=1}^{n_i} \hat{z}_i^j + v_i^j, \quad i=0,\ldots,k.
\end{equation}
In Eq.~\eqref{eq:Z_i}, $z_i^j = [u_i^j, v_i^j]^{\top}$, $\hat{z}_i^j = [\hat{u}_i^j, \hat{v}_i^j]^{\top}$, and $u_i^j, v_i^j$ correspond to the pixel measurements of the $j$-th scene landmark $\mathbf{l}_j^{\mathcal{T}}= [l_{j,x}^{\mathcal{T}}, l_{j,y}^{\mathcal{T}}, l_{j,z}^{\mathcal{T}}]^{\top}$ observed from pose $T_i$.
In this study, landmarks represent three-dimensional feature points corresponding to a set of image keypoints tracked across different camera frames.
The camera measurement noise $v_i^j$ is modelled as a zero-mean Gaussian process, $v_i^j \sim \mathcal{N} (0_{2\times 1}, \Sigma_{\text{V}})$.

The relationship between $T_i$, $\mathbf{l}_j$ and $\hat{z}_i^j$ can be derived from the pinhole camera model \cite{Szeliski2022}, 
\begin{equation}
\label{eq:PinholeCam}
    \lambda
    \begin{bmatrix}
        \hat{u}_i^j \\
        \hat{v}_i^j \\
        1
    \end{bmatrix} = K
    \begin{bmatrix}
        R_i | \mathbf{r}_i
    \end{bmatrix}
    \begin{bmatrix}
        \mathbf{l}_j^{\mathcal{T}} \\
        1
    \end{bmatrix},
\end{equation}
where $\lambda$ is the scaling parameter characterizing the projective transformation in Eq.~\eqref{eq:PinholeCam}.
Within this work, the scale ambiguity posed by $\lambda$ is resolved by assuming prior knowledge of the first two poses of the spacecraft, as will be detailed in the following.
In Eq.~\eqref{eq:PinholeCam}, the camera intrinsic matrix $K$ is composed upon knowledge of the camera focal lengths $f_x$, $f_y$ and of the principal point $c=[c_x, c_y]^{\top}$, all expressed in pixel units as follows
\begin{equation}
\label{eq:intrinsic_matrix}
    K = 
    \begin{bmatrix}
        f_x & 0 & c_x \\
        0 & f_y & c_y \\
        0 & 0 & 1
    \end{bmatrix}.
\end{equation}
Equation \eqref{eq:PinholeCam} can be further rearranged, leading to the following measurement model,
\begin{equation}
\label{eq:meas_model}
    \begin{bmatrix}
        \hat{u}_i^j \\
        \hat{v}_i^j
    \end{bmatrix} =
    \begin{bmatrix}
        f_x l_{j,x}^{\mathcal{C}} / l_{j,z}^{\mathcal{C}} + c_x \\
        f_y l_{j,y}^{\mathcal{C}} / l_{j,z}^{\mathcal{C}} + c_y
    \end{bmatrix},
\end{equation}
where $\mathbf{l}_j^{\mathcal{C}} = [l_{j,x}^{\mathcal{C}}, l_{j,y}^{\mathcal{C}}, l_{j,z}^{\mathcal{C}}]^{\top}$ is the position of landmark $\mathbf{l}_j$ in camera frame coordinates,
\begin{equation}
    l_j^{\mathcal{C}} = 
    \begin{bmatrix}
        R_i | \mathbf{r}_i
    \end{bmatrix}
    \begin{bmatrix}
        \mathbf{l}_j^{\mathcal{T}} \\
        1
    \end{bmatrix},
\end{equation}
and we have assumed that $l_{j,z}^{\mathcal{C}} \neq 0$.

The general factor graph-based SLAM framework can be specialized to estimate the spacecraft pose trajectory and the triangulated scene landmarks based on the monocular measurements $Z_{0:k}$.
An augmented state $X_k = \{T_{0:k}, M_k \}$ collecting poses $T_{0:k}$ and map variables $M_k$ is introduced, whose posterior distribution $p(X_k | Z_{0:k})$ is referred to as the \textit{belief} $\beta(X_k)$, given by
\begin{equation}
    \beta(X_k) \triangleq p(X_k | Z_{0:k}).
\end{equation}
The belief
$\beta(X_k)$ has a factorization in the form of Eq.~\eqref{eq:factor_graph_pdf}, in which factors $\phi_i^j$ are divided into prior and projection factors, $\phi_i^{\text{prior}}$ and $\phi_i^{j, \text{proj}}$, respectively.
Both factor types can be seen as a specialized version of Eq.~\eqref{eq:general_factor},
\begin{align}
    \phi_i^{\text{prior}} &\triangleq \exp \left( -\frac{1}{2} \lVert T_i^{\text{prior}} - T_i \rVert^2_{\Sigma_{\text{P}}} \right), \\
    \phi_i^{j, \text{proj}} &\triangleq \exp \left( -\frac{1}{2} \lVert \hat{z}_i^j - z_i^j \rVert^2_{\Sigma_{\text{V}}} \right),
\end{align}
where $\lVert \cdot \rVert^2_{\Sigma}$ is the squared Mahalanobis norm and $T_i^{\text{prior}}$ is a known value inferred from prior knowledge of the problem.
Hence, the following factorization applies,
\begin{equation}
\label{eq:belief_state}
    \beta(X_k) \propto \phi_0^{\text{prior}} \phi_1^{\text{prior}} \prod_{i=0}^k \prod_{j=1}^{n_i} \phi_i^{j, \text{proj}},
\end{equation}
where $\phi_0^{\text{prior}}$ and $\phi_1^{\text{prior}}$ are priors on the first two poses $T_0$ and $T_1$.
Recalling Eq.~\eqref{eq:factor_graph_MAP_1}, the SLAM solution can be extracted from Eq.~\eqref{eq:belief_state} in terms of the MAP estimate, $X_k^{\text{MAP}} = \{T_{0:k}^{\text{MAP}}, M_k^{\text{MAP}} \}$.

Equation \eqref{eq:belief_state} describes the belief state at $t_k$ predicated on a sequence of past observations and on prior hypothesis on the distribution of the first two poses.
The same approach can be adopted to predict a belief $\beta(X_{k+L})$ at a future time step $t_{k+L}$.
In fact, if the agent maintains a representation of the world and a model of its motion, the measurement function in Eq.~\eqref{eq:meas_model} can be employed to predict future observations $\hat{Z}_{k+1:k+L}=\{\hat{Z}_{k+1}, \ldots, \hat{Z}_{k+L} \}$.
After constructing the projection factors corresponding to the predicted measurements, the future belief $\beta(X_{k+L})$ can be represented explicitly as 
\begin{equation}
\label{eq:future_bel_state}
    \beta(X_{k+L}) \propto \beta(X_k) \prod_{i=k+1}^{k+L} \prod_{j=1}^{n_i} \phi_i^{j, \text{proj}},
\end{equation}
where $\beta(X_k)$ follows from Eq.~\eqref{eq:belief_state}.

Building upon Eq.~\eqref{eq:future_bel_state}, information-theoretic active SLAM can be formally stated as the problem of identifying a reference pose trajectory $T^*_{k+1 : k+L}$ such that
\begin{equation}
\label{eq:active_SLAM}
    T^*_{k+1 : k+L} = \argmax_{T_{k+1 : k+L}} \Biggl\{ \mathop{\mathbb{E}}_{Z_{k+1:k+L}} \Bigg[\sum_{i=k+1}^{k+L-1} \mathcal{L}\left( \beta(X_i)\right)
    + \mathcal{Q} \left(\beta(X_{k+L}) \right) \Bigg] \Biggr\},
\end{equation}
where both the stage reward $\mathcal{L}\left( \beta(X_i) \right)$ and the terminal reward $\mathcal{Q} \left(\beta(X_{k+L}) \right)$ encode a measure of the information associated to a specific belief state $\beta(X)$.

\section{Solution Approach} \label{sec:approach}

\subsection{Solution Outline}
\label{sec:sol_outline}

Different assumptions \cite{Placed2023survey} on the structure of the problem in Eq.~\eqref{eq:active_SLAM} can be made to achieve a tractable solution. 
We neglect the stage rewards and only consider the terminal $\mathcal{Q} \left(\beta(X_{k+L}) \right)$. \cite{Kopitkov2017bsp}
While, in principle, $\mathcal{Q} \left(\beta(X_{k+L}) \right)$ might include contributions distinct from the information-related ones (e.g., distance from a desired goal state), in our analysis we focus on information gain as the terminal reward,
\begin{equation}
\label{eq:info_gain}
    \mathcal{Q}\left( \beta(X_{k+L}) \right) \triangleq \mathcal{H}\left( \beta(X_{k}) \right) - \mathcal{H}\left( \beta(X_{k+L}) \right),
\end{equation}
where $\mathcal{H}\left( \beta(X) \right)$ is the differential entropy associated to the prior belief $\beta(X)$, i.e., the belief prior to measurements $\hat{Z}_{k+1:k+L}$.
The reward function in Eq.~\eqref{eq:info_gain} captures the reduction in uncertainty (or, equivalently, the amount of information) resulting from a certain course of action.
When the density $\beta(X)$ is an $n$-dimensional Gaussian with covariance $\Sigma$, its entropy has the following expression, 
\begin{equation}
    \label{eq:gaussian_entropy}
    \mathcal{H}\left( \beta(X) \right) = \frac{n}{2} \log(2\pi e) + \frac{1}{2}\log \lvert \Sigma \rvert.
\end{equation}
Substituting Eq.~\eqref{eq:gaussian_entropy} into Eq.~\eqref{eq:info_gain} and using the information matrix $\Lambda=\Sigma^{-1}$ leads to a closed-form expression of the information gain,
\begin{equation}
\label{eq:info_gain_1}
    \mathcal{Q}\left( \beta(X_{k+L}) \right) = -\frac{n^{\prime}}{2}\log(2\pi e) + \frac{1}{2} \log  \frac{\lvert \Lambda_{k+L} \rvert}{\lvert \Lambda_k \rvert},
\end{equation}
where $n^{\prime}=\textrm{dim}(X_{k+L}) - \textrm{dim}(X_{k})$.
Note that, for an active SLAM setting, $\textrm{dim}(X_{k+L})>\textrm{dim}(X_{k})$ since the augmented state $X_{k+L}$ includes future poses.
The computation of the determinants in Eq.~\eqref{eq:info_gain_1} is, in general, $\mathcal{O}(n^3)$, which hinders real-time planning when the problem is high-dimensional.
However, methods exist to decouple the time complexity from the state dimensionality, reducing the cost of evaluating the information gain to a function of the planning horizon $L$. \cite{Kopitkov2017bsp}

Given a current belief $\beta(X_k)$, we consider a collection $\mathcal{S}$ of $M$ candidate pose trajectories along the planning horizon $t_{k+1:k+L}$,
\begin{equation}
\label{eq:C_set}
    \mathcal{S} = \{T^1, \ldots, T^M \},
\end{equation}
with $T^m=T_{k+1:k+L}^m$, $m=1,\ldots,M$.
Since the belief depends on the state trajectory, recalling Eq.~\eqref{eq:future_bel_state} a set $\mathcal{B}$ consisting of $M$ future beliefs resulting from the candidate actions in Eq.~\eqref{eq:C_set} can be constructed,
\begin{equation}
\label{eq:B_set}
    \mathcal{B} = \{\beta^1(X_{k+L}), \ldots, \beta^M(X_{k+L}) \}.
\end{equation}
Then, a score $r^m$ can be assigned to each $\beta^m(X_{k+L})\in \mathcal{B}$,
\begin{equation}
\label{eq:reward_m_1}
    r^m = \mathop{\mathbb{E}}_{Z_{k+1:k+L}} \Bigg[ \mathcal{Q}\left(\beta^m(X_{k+L}) \right) \Bigg] 
    = \mathop{\mathbb{E}}_{Z_{k+1:k+L}} \Bigg[ 
    -\frac{n^{\prime}}{2}\log(2\pi e) + \frac{1}{2} \log  \frac{\lvert \Lambda_{k+L}^m \rvert}{\lvert \Lambda_k \rvert}
    \Bigg]
\end{equation}
hence obtaining a reward set $\mathcal{R}$,
\begin{equation}
\label{eq:reward_set}
    \mathcal{R} = \{r^1, \ldots, r^M\}.
\end{equation}

\subsection{Active Satellite SLAM}
\label{sec:solution}

We now tailor the steps presented in the previous section to the spacecraft proximity scenario of interest.
First, we limit our investigation to the case where no position control is applied to the chaser spacecraft, as described by Eqs.~\eqref{eq:CWx}-\eqref{eq:CWz}.
This assumption reflects a realistic close-proximity inspection phase, where the inspecting spacecraft must minimize thruster fuel consumption, relying solely on attitude control to plan informative maneuvers.

Secondly, in optimizing the chaser's attitude profile, we restrict the optimization to attitude trajectories where the camera's boresight remains directed towards a fixed point in the scene along the chaser's motion. 
This approach aligns the complexity of our attitude trajectory with the ``center-pointing'' strategy commonly used in passive navigation solutions. \cite{Opromolla2015UncoopPose, Elankumaran2021Distributed, dor2022astroslam} 
A notable implication of this choice is that our method, in principle, does not demand higher attitude tracking performance than what is typically required in passive scenarios.

Given the predicted spacecraft path $\hat{\mathbf{r}}_{k+1:k+L}$ and velocity $\hat{\mathbf{v}}_{k+1:k+L}$ obtained by solving Eqs.~\eqref{eq:CWx}-\eqref{eq:CWz}  with $\mathbf{w}_C^{\mathcal{T}}=0_{3\times 1}$, a candidate attitude trajectory $R_{k+1:k+L}^m$ can be computed using Eq.~\eqref{eq:c123} by simply selecting an observation target $\mathbf{r}_{OT}^{\mathcal{T},m}$ ($\mathbf{r}_O^m$ for brevity).
Since the pose trajectory candidates $T_{k+1:k+L}^m$ introduced in Eq.~\eqref{eq:C_set} are fully determined by the choice of $\mathbf{r}_O^m$, we recall Eq.~\eqref{eq:reward_m_1} and rewrite the active SLAM problem in Eq.~\eqref{eq:active_SLAM} in the following way,
\begin{equation}
\label{eq:active_SLAM_1}
        \mathbf{r}_O^* = \argmax_{\mathbf{r}_O}
     \Biggl\{ \mathop{\mathbb{E}}_{Z_{k+1:k+L}} \Bigg[ -\frac{n^{\prime}}{2}\log(2\pi e) 
     + \frac{1}{2} \log  \frac{\lvert \Lambda_{k+L}^m \rvert}{\lvert \Lambda_k \rvert} \Bigg] \Biggr\}.
\end{equation}

\begin{figure}[t]
\centering
\includegraphics[width=3.3in]{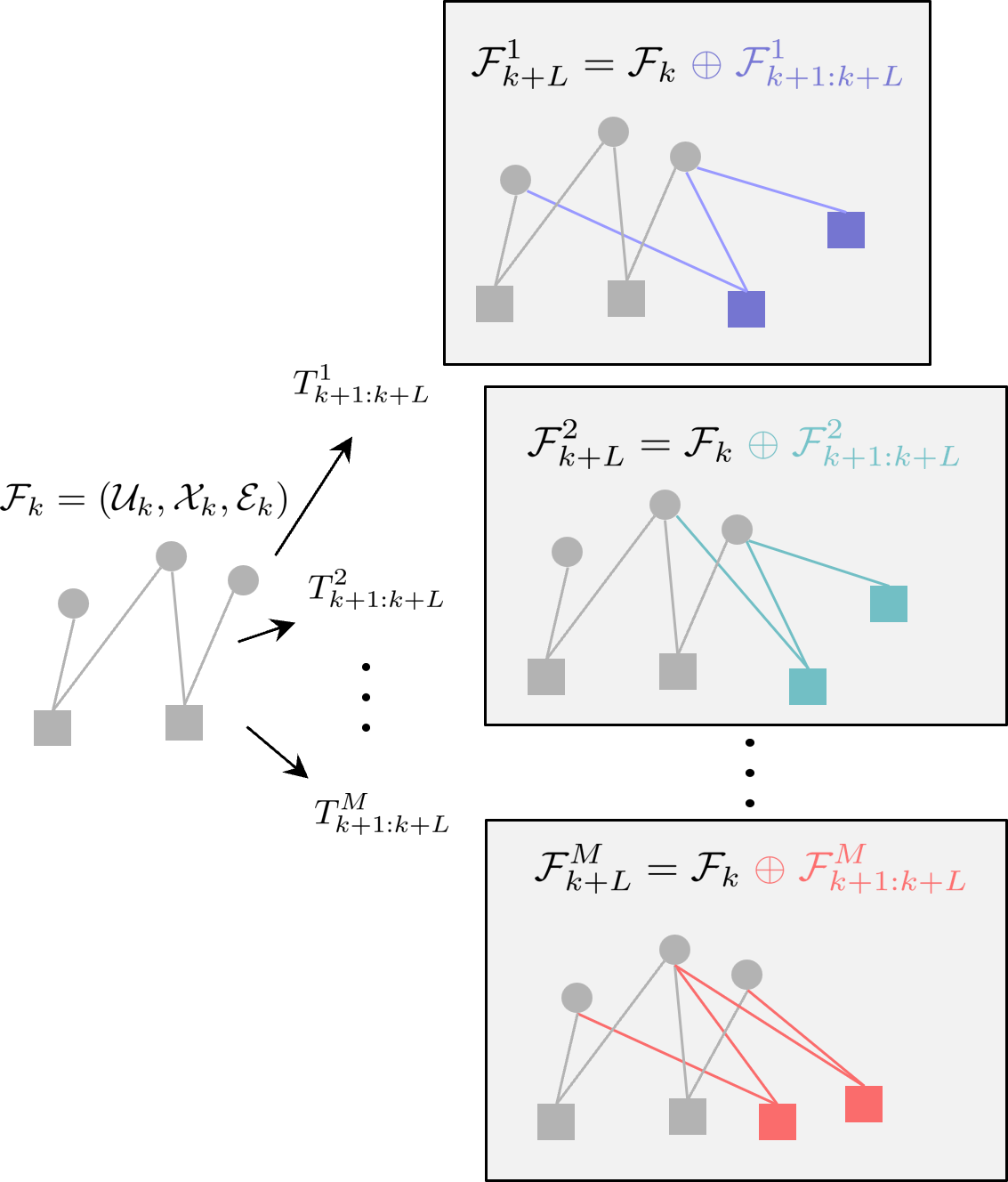}
\caption{Schematic of the graph augmentation operations. 
$\mathcal{F}_k$ is the factor graph modeling the current belief $\beta(X_k)$, from which the posterior $\beta^m(X_{k+L})$ can be obtained by ``augmenting'' $\mathcal{F}_k$ with $\mathcal{F}^m_{k+1:k+L}$.
Squares and circles represent poses and map variables, respectively, edges encode probabilistic constraints (factors) arising from observations.
Future poses and factors in color.
}
\label{fig:graph_augm}
\end{figure}

The evaluation of the posterior information matrices $\Lambda_{k+L}^m$ in Eq.~\eqref{eq:active_SLAM_1} follows from a graph augmentation operation which we describe with the symbol $\oplus$.
For each $m$-th candidate plan, an ``augmented'' factor graph $\mathcal{F}^m_{k+L} = \mathcal{F}_k \oplus \mathcal{F}^m_{k+1:k+L}$ modeling the posterior $\beta^m(X_{k+L})$ can be formed by augmenting the current graph $\mathcal{F}_k$ with $\mathcal{F}^m_{k+1:k+L}$, the graph encoding the probabilistic relationships between future poses and current map variables (see Fig.~\ref{fig:graph_augm}); formally,

\begin{equation}
\label{eq:graph_new}
    \mathcal{F}^m_{k+1:k+L} = \left(\mathcal{U}^m_{k+1:k+L}, \mathcal{X}^m_{k+1:k+L}, \mathcal{E}^m_{k+1:k+L} \right).
\end{equation}
In Eq.~\eqref{eq:graph_new}, factors $\mathcal{U}^m_{k+1:k+L}$ stem from predicted observations of already mapped landmarks,
\begin{equation}
\label{eq:new_factors}
    \mathcal{U}^m_{k+1:k+L} = \bigcup_{i=k+1}^{k+L} \bigcup_{j=1}^{n_i} \prescript{m}{}\phi_i^{j, \textrm{proj}},
\end{equation}
while variables $\mathcal{X}^m_{k+1:k+L}$ include future poses and observed landmarks $M_{k+L}^m$,
\begin{equation}
\label{eq:new_variables}
    \mathcal{X}^m_{k+1:k+L} = T_{k+1:k+L}^m \cup M_{k+L}^m,
\end{equation}
and edges $\mathcal{E}^m_{k+1:k+L}$ connect members of $\mathcal{U}^m_{k+1:k+L}$ and $\mathcal{X}^m_{k+1:k+L}$.
Note that, while new poses and factors are introduced in each $\mathcal{F}^m_{k+1:k+L}$, the map variables in $\mathcal{X}^m_{k+1:k+L}$ correspond to landmarks already present in $\mathcal{F}_k$. 
This is because the structure of $\mathcal{F}_{k+1:k+L}^m$ is predicated only on the current knowledge of the environment.
Estimating the information level associated to non-mapped regions of the scene is out of the scope of this work and constitutes a possible further development.

Once the augmented graphs $\mathcal{F}^m_{k+1:k+L}$, $m=1,\ldots,M$ are available, each corresponding to a specific camera pointing strategy $\mathbf{r}_O^m$, the measurement Jacobians $A_{k+L}^m$ can be obtained upon linearization of the nonlinear projection factors in $\mathcal{U}_{k+1:k+L}^m$.
From the Jacobians, the information matrices $\Lambda_{k+L}^m$ necessary to evaluate the determinants in $r^m$ (Eq.~\eqref{eq:reward_m_1}) can be obtained as follows,
\begin{equation}
\label{eq:At_A}
    \Lambda_{k+L}^m = (A_{k+L}^m)^{\top}A_{k+L}^m.
\end{equation}
In a similar fashion, the determinant of the information matrix $\Lambda_k$ associated with $\beta(X_k)$ can be computed and stored at the beginning of the procedure.
Finally, $\mathbf{r}_O^*$ is found by constructing $\mathcal{R}$ as in Eq.~\eqref{eq:reward_set} and associating $r^*=\max(\mathcal{R})$ to the corresponding $\mathbf{r}_O^*$.
The steps of the procedure above are reported in Alg.~\ref{alg:active_SatSLAM}.

 \begin{algorithm}[ht]
 \caption{Active Satellite SLAM}
 \begin{algorithmic}[1]
 \label{alg:active_SatSLAM}
 \renewcommand{\algorithmicrequire}{\textbf{Input:}}
 \renewcommand{\algorithmicensure}{\textbf{Output:}}
 \REQUIRE Nonlinear factor graph $\mathcal{F}_k$, number of candidate trajectories $M$
 \ENSURE  Informative observation goal $\mathbf{r}_O^*$
  \STATE Linearize $\mathcal{F}_k$ around current guess $\hat{\mathcal{X}}_k$
  \STATE Compute and store $\Lambda_k$
  \STATE Sample $M$ candidate observation targets $\mathbf{r}_O^m$, propagate corresponding pose trajectories $\hat{T}_{k+1:k+L}^m$ \label{lst:line:sample}
  \FOR {$m = 1$ to $M$}
  \STATE Construct $\mathcal{F}_{k+1:k+L}^m$ (Eqs.~\eqref{eq:graph_new}-\eqref{eq:new_variables})
  \STATE $\mathcal{F}_{k+L}^m=\mathcal{F}_k \oplus \mathcal{F}_{k+1:k+L}^m$ (Fig.~\ref{fig:graph_augm})
  \STATE Linearize $\mathcal{F}_{k+L}^m$ around $\mathcal{X}_{k+L}^m$ and compute $\Lambda_{k+L}^m$ (Eq.~\eqref{eq:At_A})
  \STATE Compute and store $r^m = \mathbb{E}_{Z_{k+1:k+L}} \left[ \mathcal{Q}(\beta^m(X_{k+L})) \right]$ (Eqs.~\eqref{eq:reward_m_1}-\eqref{eq:reward_set})
  \ENDFOR
  \STATE Compute $r^* = \max(\mathcal{R})$
  \STATE Associate $\mathbf{r}_O^*$ to $r^*$
 \RETURN $\mathbf{r}_O^*$
 \end{algorithmic}
 \end{algorithm}

\section{Experimental Results} 
\label{sec:experiments}

\subsection{Simulation Setup}
\label{sec:sim_setup}

In this section, we detail the simulation setup designed to validate our factor graph-based active SLAM approach, along with the numerical results obtained. 
\begin{table}[!t]
\renewcommand{\arraystretch}{0.85}
\caption{Camera Parameters}
\label{tab:table_example}
\centering
\begin{tabular}{|c|c|}
\hline
\textbf{\textsc{Parameter}} & \textbf{\textsc{Value}} - pixel units\\
\hline
$f_x$ & $256$\\
\hline
$f_y$ & $256$\\
\hline
Resolution & $512\times512$ \\
\hline
$c_x$ & $256$\\
\hline
$c_y$ & $256$\\
\hline
Noise $\sigma$ & $2$ \\
\hline
\end{tabular}
\end{table}
\begin{figure*}[!t]
\centering
\begin{minipage}{\textwidth}
\centering
\subfloat[]{\includegraphics[width=1.25in]{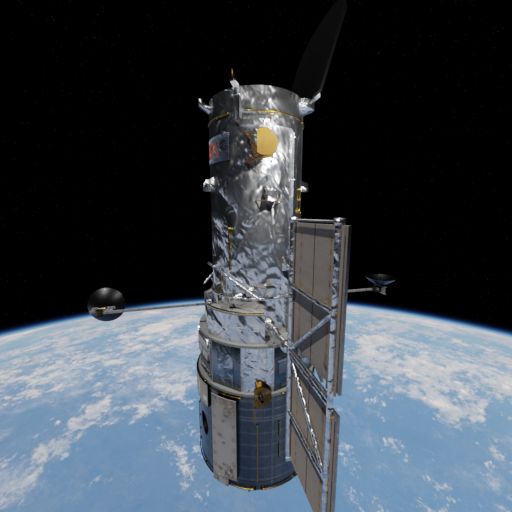}\label{fig:hst_rgb_1}}
\hfil
\subfloat[]{\includegraphics[width=1.25in]{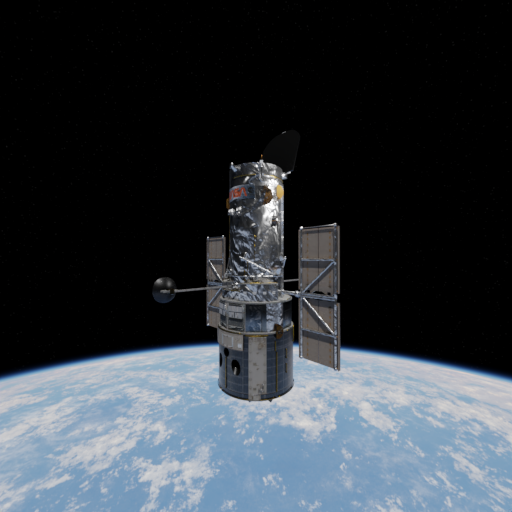}\label{fig:hst_rgb_2}}
\hfil
\subfloat[]{\includegraphics[width=1.25in]{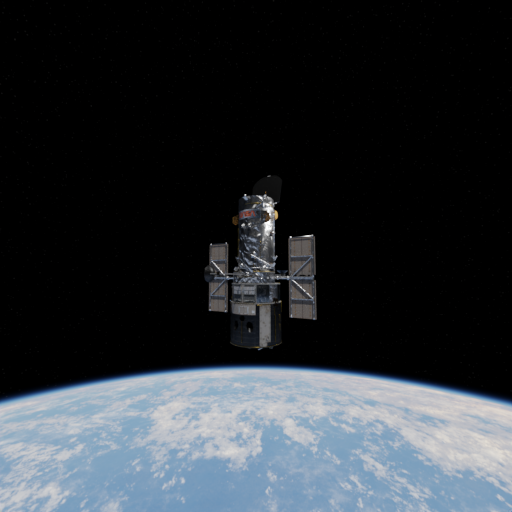}\label{fig:hst_rgb_3}}
\hfil
\subfloat[]{\includegraphics[width=1.25in]{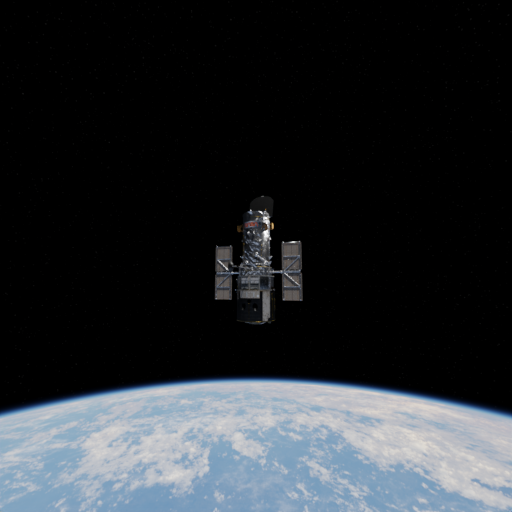}\label{fig:hst_rgb_4}}
\end{minipage}
\begin{minipage}{\textwidth}
\centering
\subfloat[]{\includegraphics[width=1.25in]{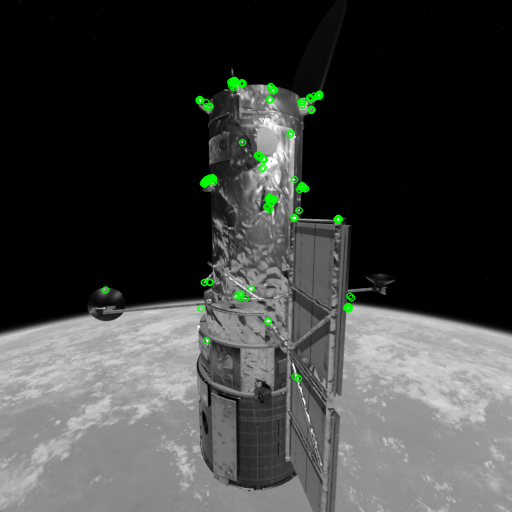}\label{fig:hst_gray_1}}
\hfil
\subfloat[]{\includegraphics[width=1.25in]{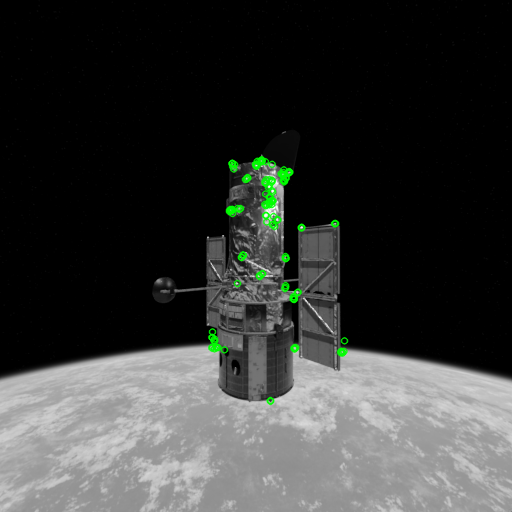}\label{fig:hst_gray_2}}
\hfil
\subfloat[]{\includegraphics[width=1.25in]{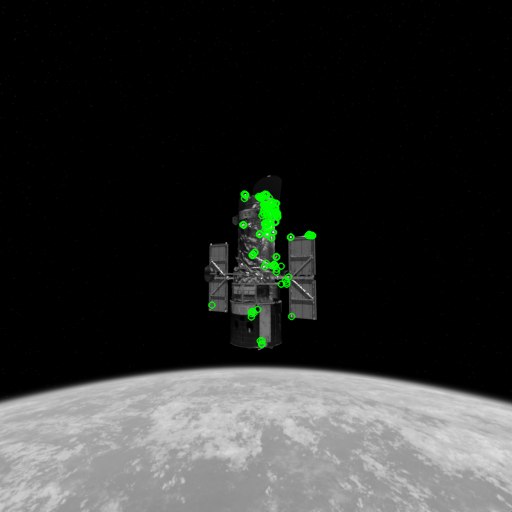}\label{fig:hst_gray_3}}
\hfil
\subfloat[]{\includegraphics[width=1.25in]{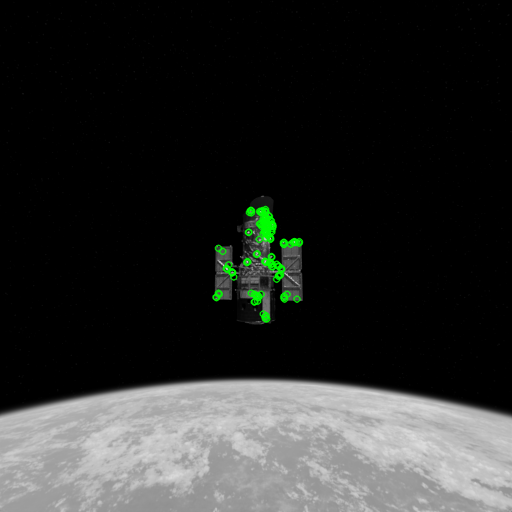}\label{fig:hst_gray_4}}
\end{minipage}
\caption{Rendered images of the HST as seen from the chaser onboard camera during the reconnaissance phase (Figs.~\ref{fig:hst_rgb_1}-\ref{fig:hst_rgb_4}) with the ORB keypoints used in the map initialization step (green dots in Figs.~\ref{fig:hst_gray_1}-\ref{fig:hst_gray_4}).}
\label{fig_sim}
\end{figure*}
\begin{table}[!t]
\renewcommand{\arraystretch}{0.85}
\caption{Orbital Parameters}
\label{tab:orbital_params}
\centering
\begin{tabular}{|c|c|c|}
\hline
\textbf{\textsc{Parameter}} & \textbf{\textsc{Symbol}} & \textbf{\textsc{Value}} \\
\hline
HST orbit altitude & $h$ & $550$ - [km] \\
\hline
Chaser initial position & $\mathbf{r}_{CT}^{\mathcal{T},0}$ & $[1,6,5]^{\top}$ - [m] \\
\hline
Chaser initial velocity & $\mathbf{v}_{CT}^{\mathcal{T},0}$ & $[0.0131, -0.0022, 0]^{\top}$ - [m/s] \\
\hline
Disturbance acceleration & $\mathbf{w}_{CT}^{\mathcal{T}}$ & $\mathcal{N}(0_{3\times 1},\textrm{1e-10}\cdot \mathbf{I}_3)$ - $\textrm{[m/s}^2\textrm{]}$ \\
\hline
\end{tabular}
\end{table}
\begin{figure*}[!t]
\centering
\begin{minipage}{\textwidth}
\centering
\subfloat[]{\includegraphics[width=3in]{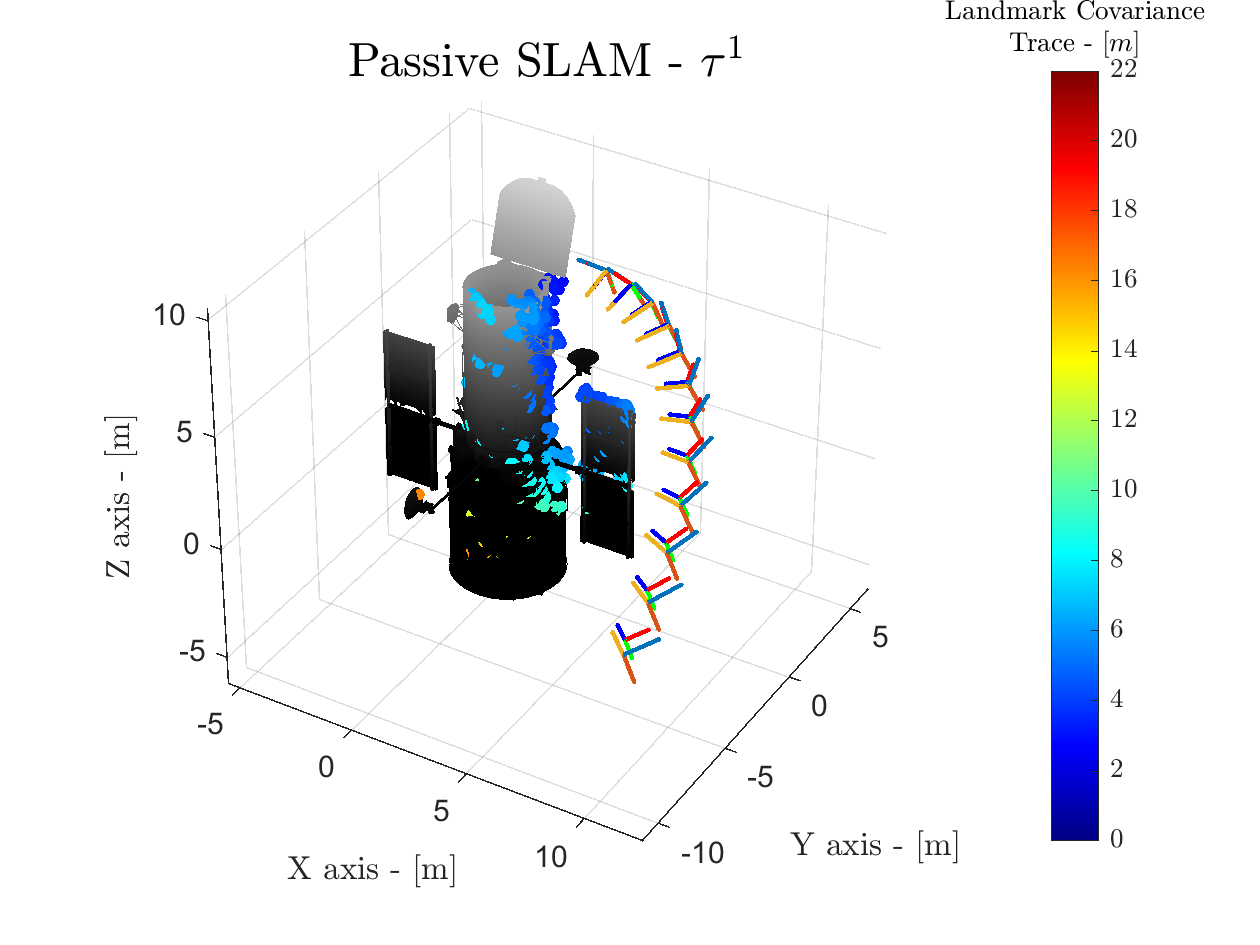}\label{fig:tau1_3D}}
\hfil
\subfloat[]{\includegraphics[width=3in]{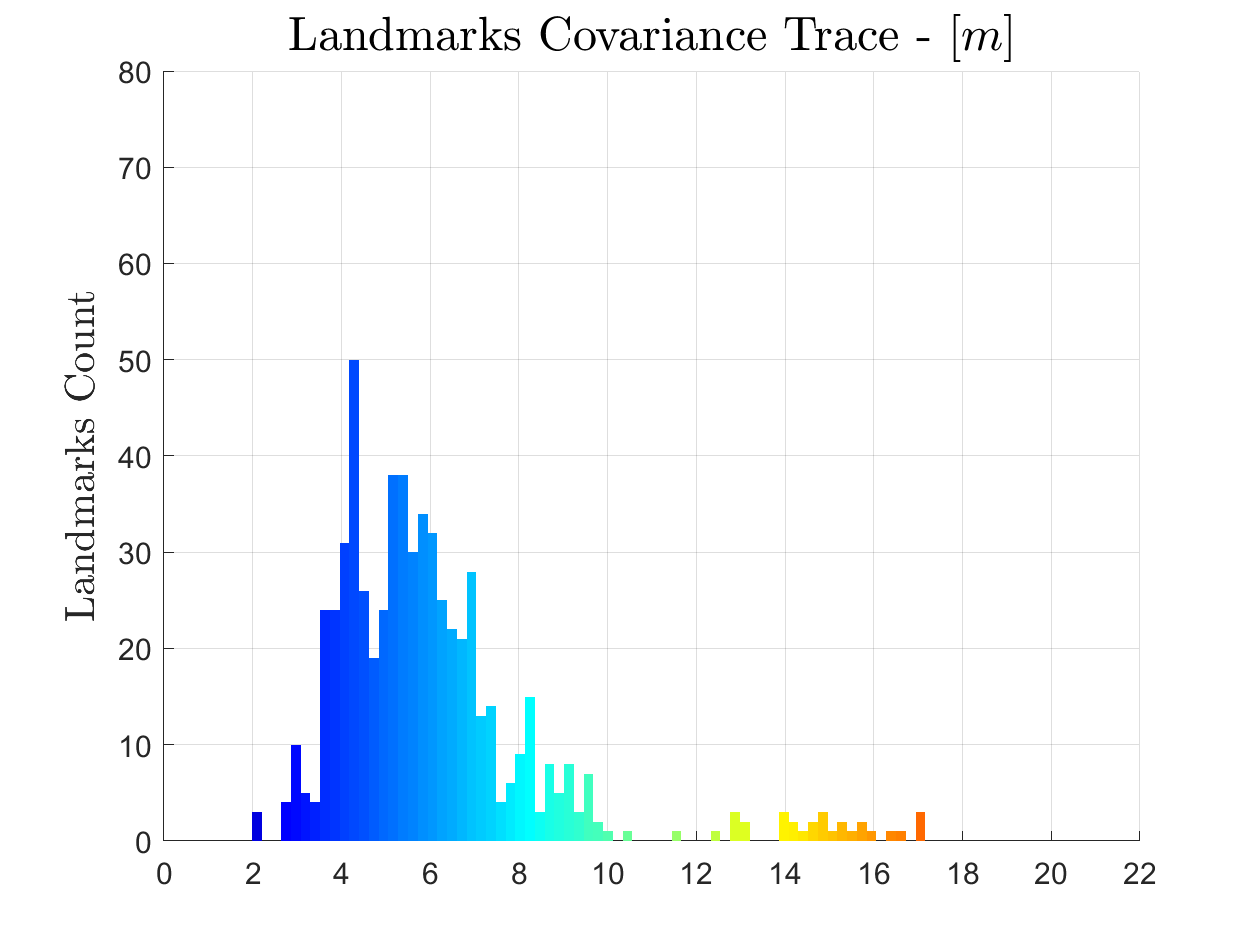}\label{fig:landm_covs_hist_tau1}}
\end{minipage}
\begin{minipage}{\textwidth}
\centering
\subfloat[]{\includegraphics[width=3in]{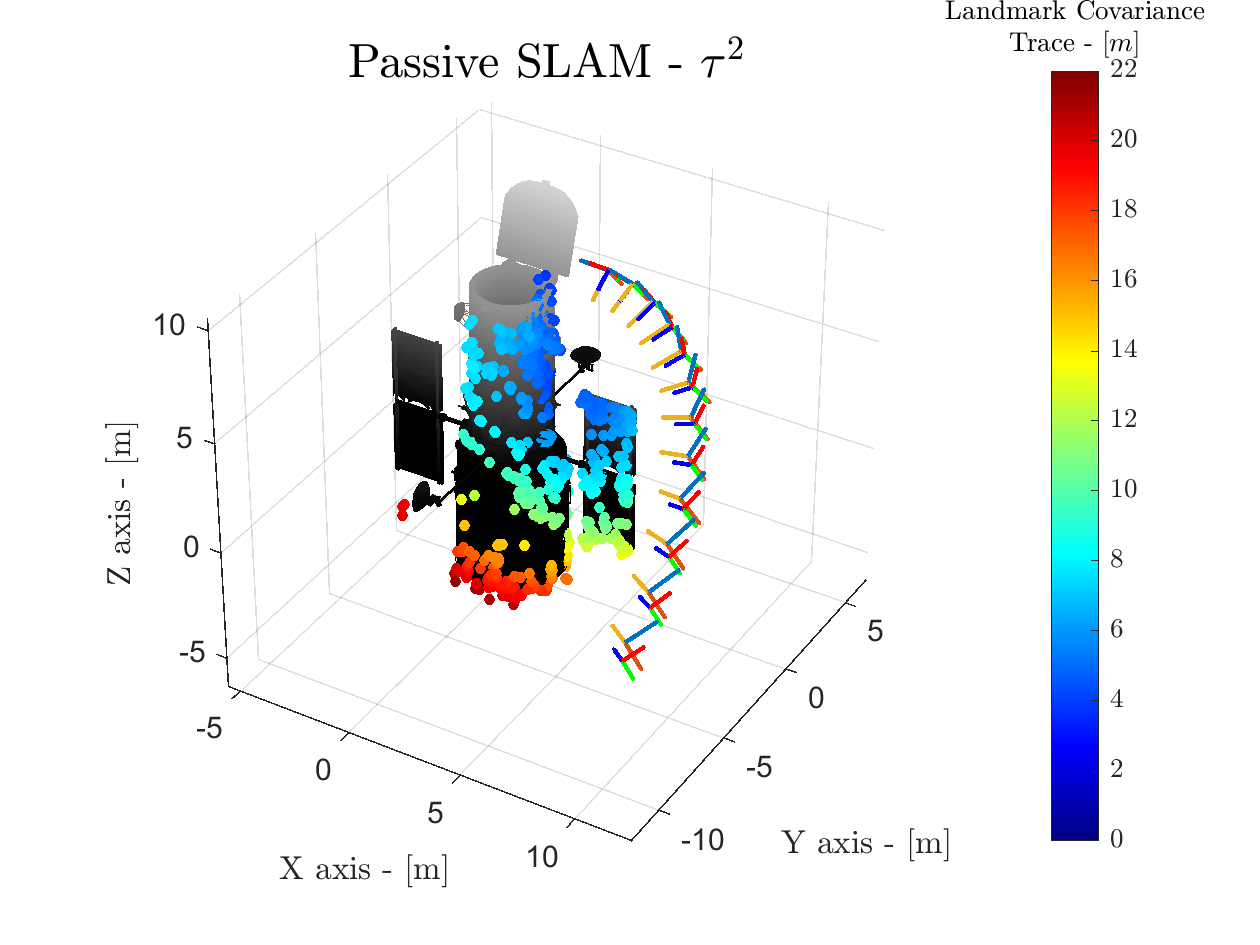}\label{fig:tau2_3D}}
\hfil
\subfloat[]{\includegraphics[width=3in]{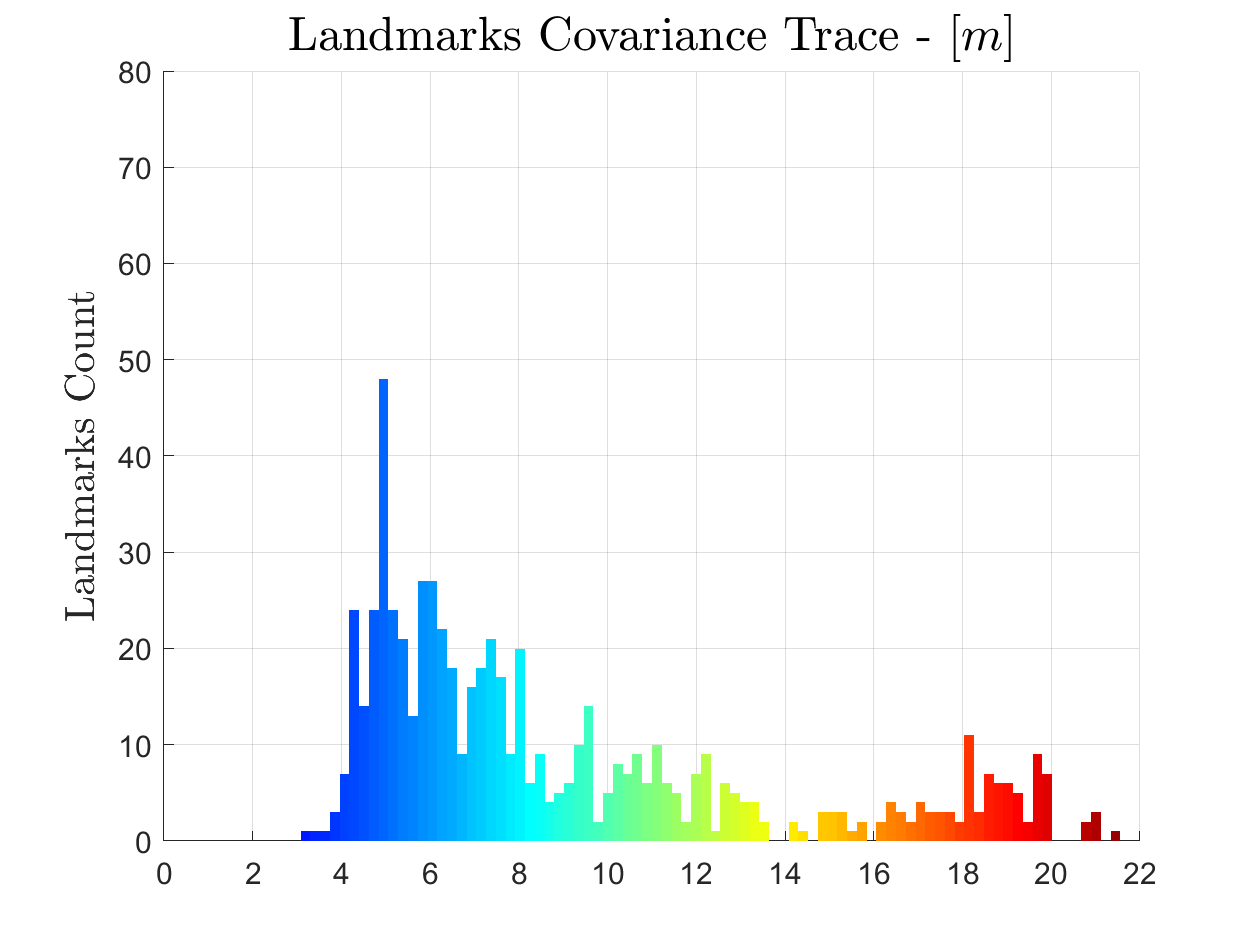}\label{fig:landm_covs_hist_tau2}}
\end{minipage}
\begin{minipage}{\textwidth}
\centering
\subfloat[]{\includegraphics[width=3in]{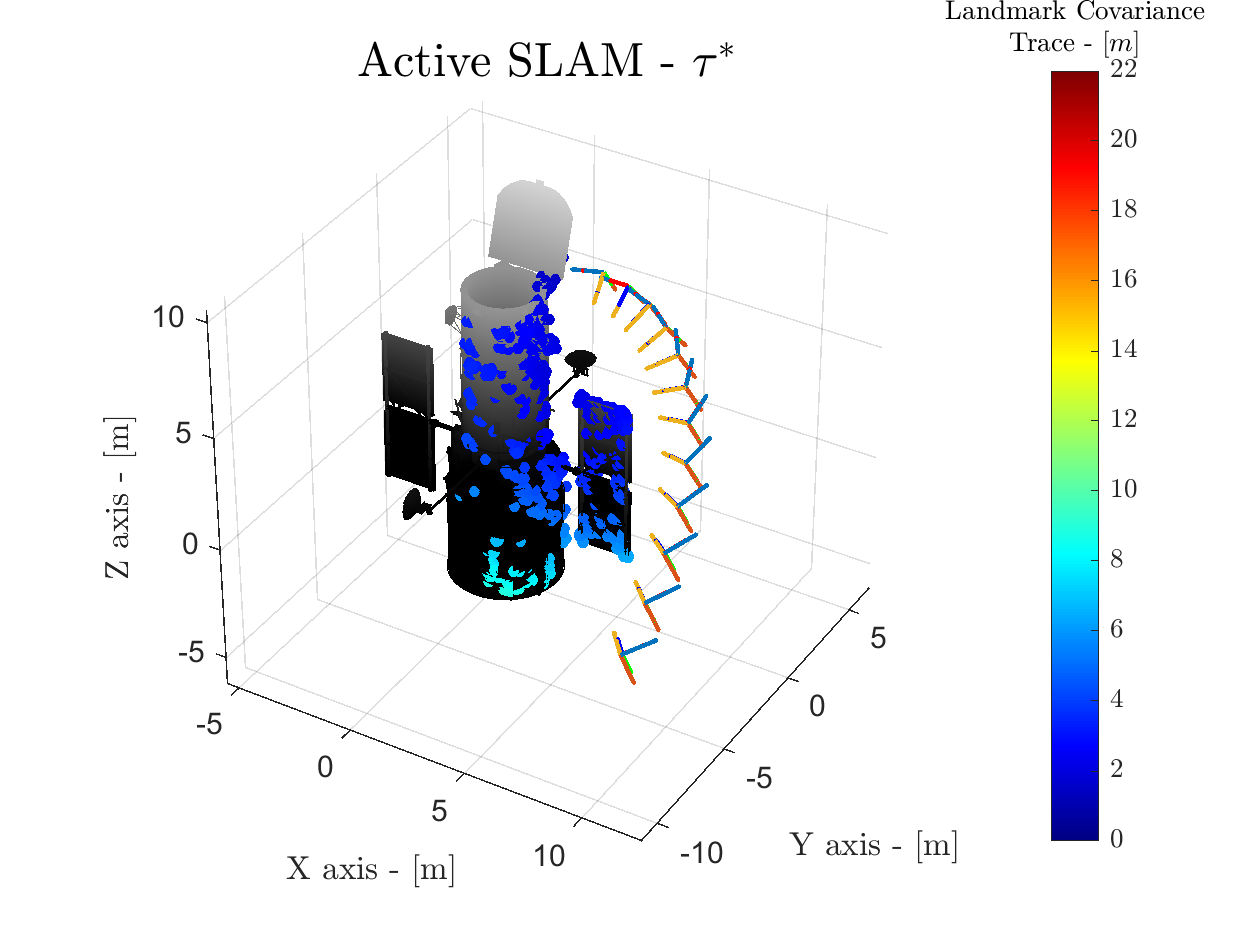}\label{fig:tau3_3D}}
\hfil
\subfloat[]{\includegraphics[width=3in]{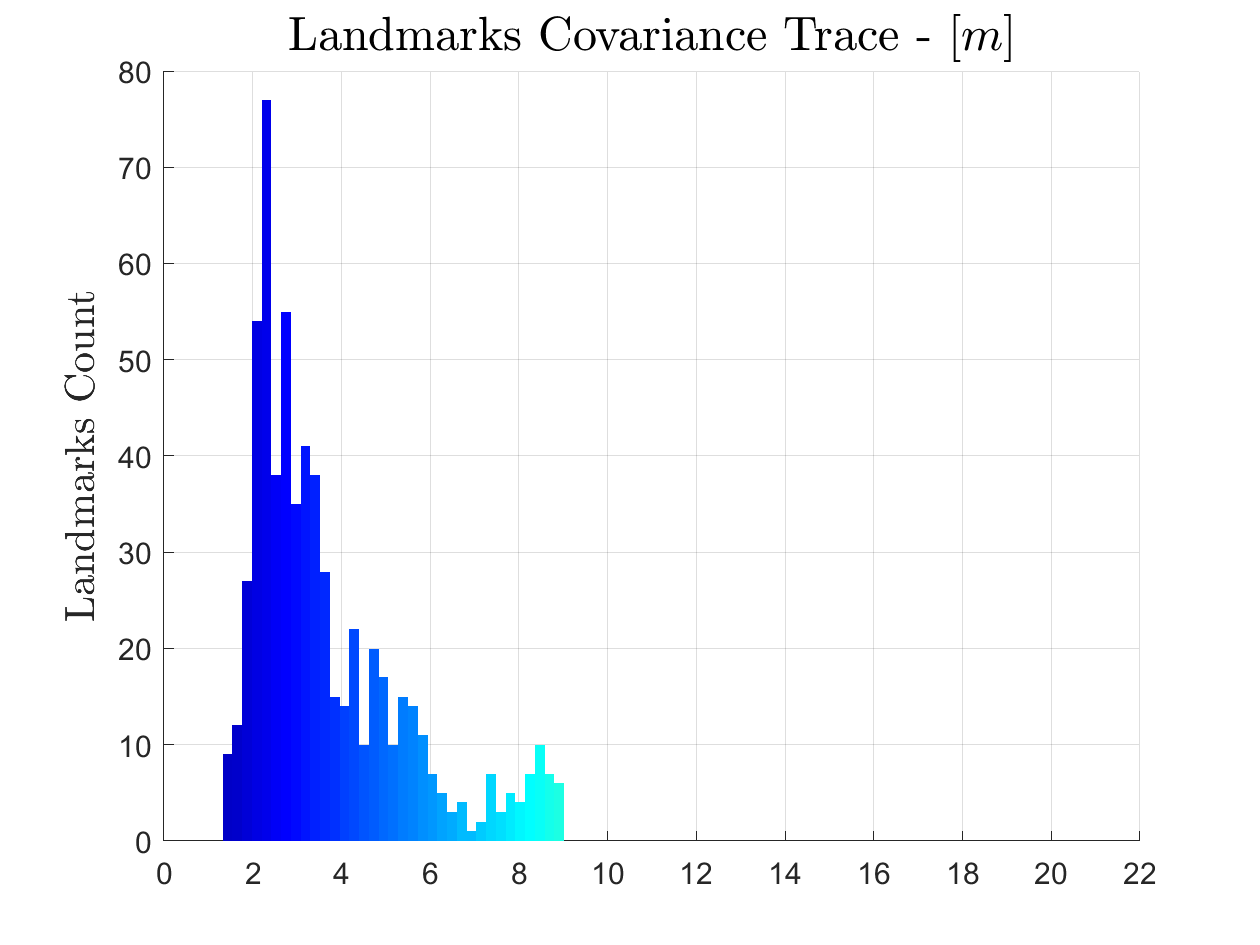}\label{fig:landm_covs_hist_taustar}}
\end{minipage}
\caption{SLAM results along trajectory $\tau^1$, $\tau^2$, $\tau^*$ for $L=L_1$. \texttt{rgb} frames represent the SLAM estimate while the \textsc{Matlab} \texttt{std} color triplet is used for the camera pose ground truth.}
\label{fig:SLAM_result}
\end{figure*}
\begin{figure}[t]
\centering
\includegraphics[width=3.4in]{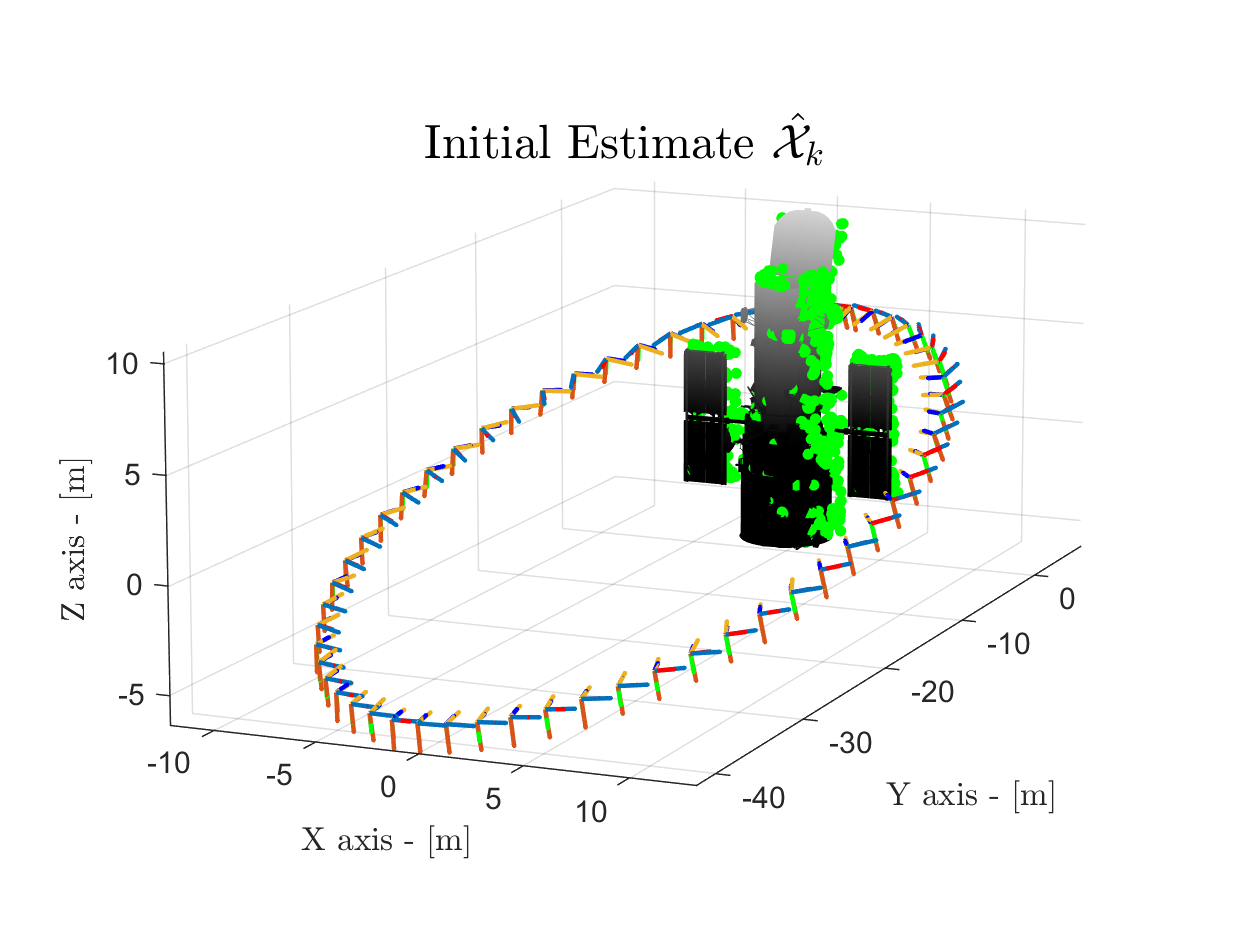}
\caption{Ground truth (colored in \textsc{Matlab} \texttt{std} triplet), estimated camera trajectory (in \texttt{rgb}) and initialized map $\hat{M}_k$ (green dots) during the reconnaissance phase around the HST.}
\label{fig:recon_phase}
\end{figure}
We simulate a proximity navigation scenario where the client spacecraft is the Hubble Space Telescope (HST) orbiting Earth in a Low Earth Orbit (LEO).
We use a 3D model of the HST available online \cite{nasaHST3d} and import it into Blender to render photorealistic high-quality images (see Figs.~\ref{fig:hst_rgb_1}-\ref{fig:hst_rgb_4}) acquired by the chaser-mounted camera characterized by the parameters in Table \ref{tab:table_example}.
Our simulations are conducted on a laptop running Ubuntu 22.04 with an Intel i7-1165G7 CPU; we use the Python bindings of the \texttt{GTSAM} library to create and manipulate factor graphs and perform graph optimization.
The synthetic camera images are processed using the \texttt{OpenCV} library, and keypoints are detected and extracted using the ORB algorithm \cite{Rublee2011orb}.

The chaser obtains an initial coarse representation of the scene, $\hat{M}_k$, from a first set of images acquired during a ``reconnaissance'' phase (Fig.~\ref{fig:recon_phase}) which consists of one relative orbit of the chaser around the HST identified by the initial conditions in Table \ref{tab:orbital_params}.
In practice, $\hat{M}_k$ is formed by a landmark point cloud, where the points are computed by back-projecting the noisy measurements from the 2D camera frame into the world frame $\mathcal{T}$ (green dots in Fig.~\ref{fig:recon_phase}).
Notably, $\hat{M}_k$ constitutes the only source of knowledge about the environment available to the planner.
The initial graph $\mathcal{F}_k$ is also constructed during the reconnaissance leg resorting to the CW equations \eqref{eq:CWx}-\eqref{eq:CWz} with $\mathbf{w}_C^{\mathcal{T}}=0_{3\times 1}$ to initialize the pose variables $\hat{T}_{0:k}$ in $\mathcal{F}_k$. 
The landmark coordinates in $\hat{M}_k$ are initialized inverting the relationship in Eq.~\eqref{eq:meas_model}, where the values $\hat{u}_i^j$ and $\hat{v}_i^j$ are the pixel coordinates of the keypoints extracted from the camera images during the reconnaissance phase (see Figs.~\ref{fig:hst_gray_1}-\ref{fig:hst_gray_4}).
No graph optimization is performed on $\mathcal{F}_k$, which is only used as an input to the active SLAM pipeline. 
An example of the solution guess $\hat{\mathcal{X}}_{0:k}=\{\hat{T}_{0:k}, \hat{M}_k\}$ obtained along the reconnaissance orbit is reported in Fig.~\ref{fig:recon_phase}.
\begin{table}[!t]
\renewcommand{\arraystretch}{0.85}
\caption{Active SLAM Parameters}
\label{tab:activeSLAM_params}
\centering
\begin{tabular}{|c|c|c|}
\hline
\textbf{\textsc{Parameter}} & \textbf{\textsc{Symbol}} & \textbf{\textsc{Value}} \\
\hline
Sampled trajectories & $M$ & $10$ \\
\hline
Lower bound for $\mathcal{D}$ & $\mathbf{l}$ & $[-1.2, -2, -2]^{\top}$ - [m]\\
\hline
Upper bound for $\mathcal{D}$ & $\mathbf{u}$ & $[2.5, 2, 5]^{\top}$ - [m]\\
\hline
Planning horizon \#1 & $L_1$ & $12$ \\
\hline
Planning horizon \#2 & $L_2$ & $23$ \\
\hline
\end{tabular}
\end{table}

The $M$ candidate observation goals evaluated by the active SLAM pipeline (see Line~\ref{lst:line:sample} in Algorithm~\ref{alg:active_SatSLAM}) are sampled from a multivariate uniform distribution $\mathcal{D}_{[\mathbf{l},\mathbf{u}]}$ around the geometric center of the HST model, with lower and upper bounds $\mathbf{l}$ and $\mathbf{u}$ defined in Table \ref{tab:activeSLAM_params}. 
The corresponding candidate attitude trajectories $\hat{R}_{k+1:k+L}^m$, $m=1,\ldots,M$, are constructed by computing the direction of the unit axes $\mathbf{c}_1, \mathbf{c}_2, \mathbf{c}_3$ in Eq.~\eqref{eq:c123} along the predicted chaser path $\hat{\mathbf{r}}_{k+1:k+L}$; 
then, the transformations $\hat{T}_{k+1:k+L}^m$ are assembled using Eq.~\eqref{eq:sc_state}.
The expectation in Eq.~\eqref{eq:reward_m_1} is resolved by inserting the predicted (expected) values of the observations $\hat{Z}_{k+1:k+L}$ in the planning graph, obtained through the measurement model in Eq.~\eqref{eq:meas_model}.

The ground truth trajectory $\mathbf{r}_{k+1:k+L}$ is simulated through numerical integration of Eqs.~\eqref{eq:CWx}-\eqref{eq:CWz} in the presence of noise, while the true landmark positions that correspond to specific features on the surface of the HST are computed from the 3D model imported in Blender.
In order to mimic the behavior of an attitude-tracking controller, the true attitude evolution $R_{k+1:k+L}$ follows the prescribed orientation strategy, on which we inject noise to reproduce possible limits and/or inaccuracies in the attitude tracking controller.
The main parameters characterizing our simulation setup are reported in Table \ref{tab:activeSLAM_params}.

In order to evaluate our planning approach without introducing additional complexities arising from other modules in the visual SLAM pipeline, we rely on a simplified visual front-end.
Therefore, we assume perfect feature matching along the simulated trajectories, hence introducing estimation errors only due to the process and measurement noise.
In order to provide an unbiased comparison, this assumption is applied when simulating both our active SLAM method and its passive counterparts, as described in the next paragraphs. 

\subsection{Performance Evaluation}
\label{sec:perf_eval}
\begin{figure*}[ht]
\centering
\begin{minipage}{\textwidth}
\centering
\subfloat[Pose and attitude errors (solid lines) and uncertainties (shaded areas) for $L=12$.]{\includegraphics[width=0.51\textwidth]{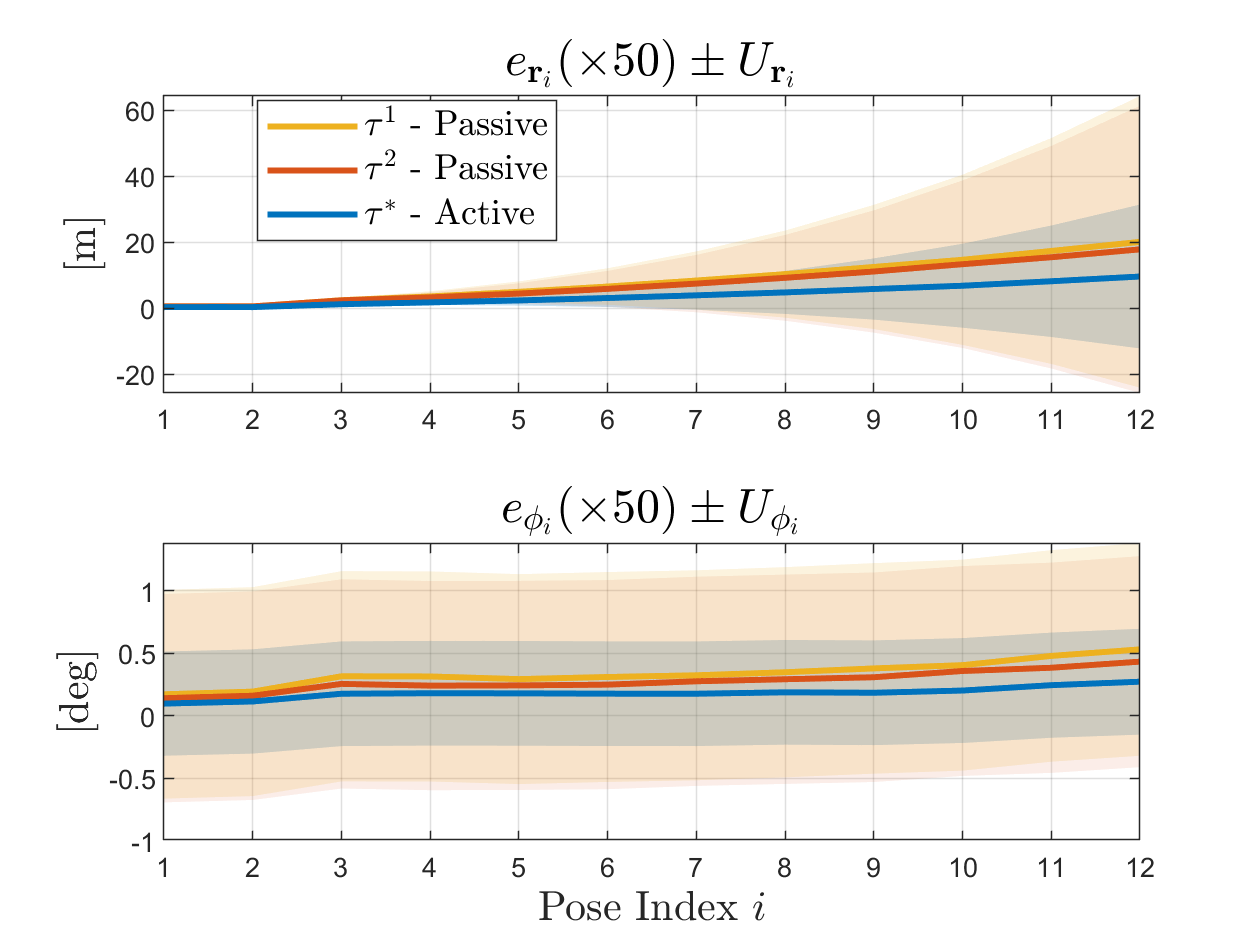}\label{fig:pos_perf_L1}}
\hfil
\subfloat[Pose and attitude errors (solid lines) and uncertainties (shaded areas) for $L=23$.]{\includegraphics[width=0.51\textwidth]{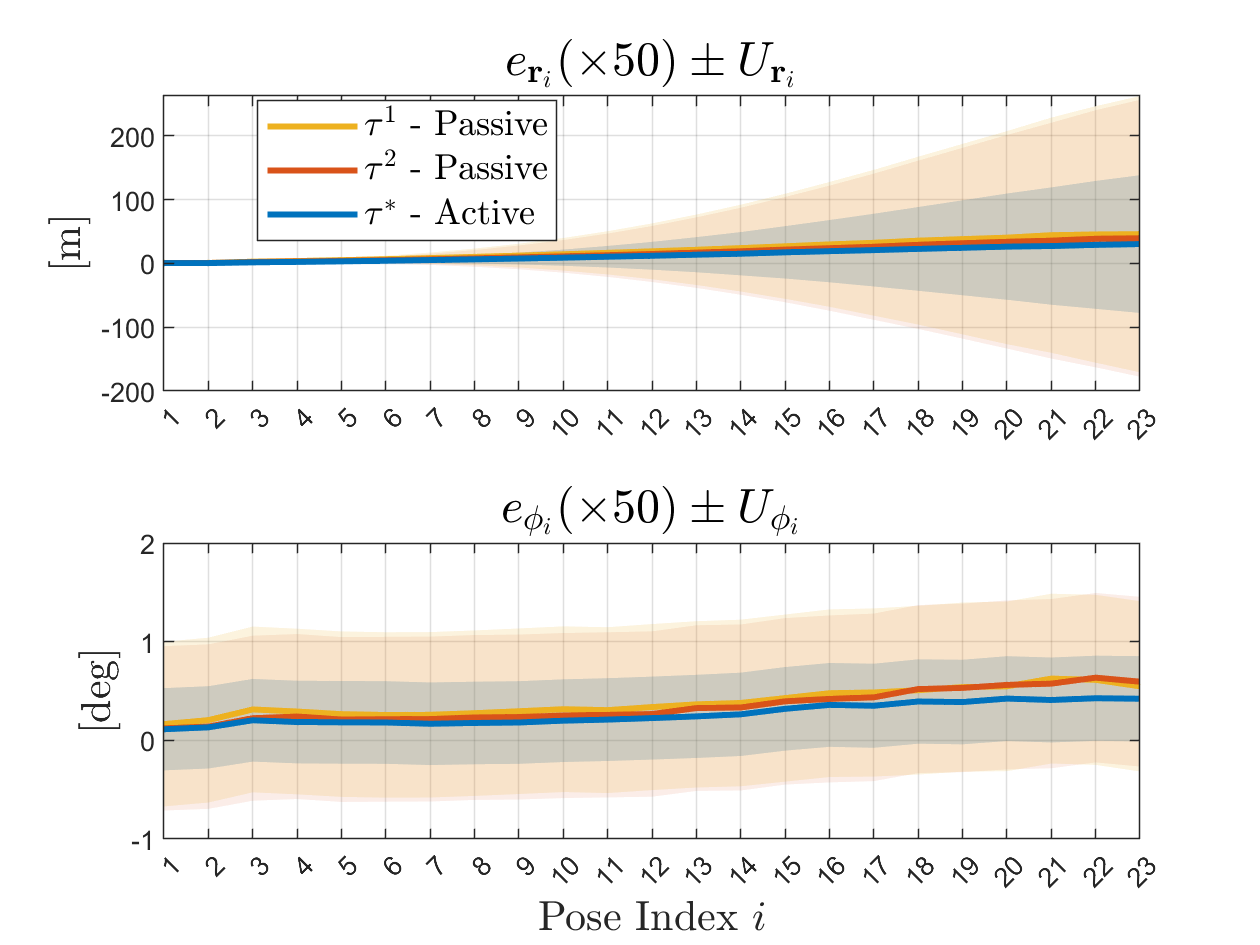}\label{fig:pos_perf_L2}}
\end{minipage}
\caption{Performance of pose trajectory estimation for time horizons $L_1$, $L_2$. The values of the errors are scaled up by a factor of $50$ for visualization purposes.}
\label{fig:pose_performance}
\end{figure*}
\begin{figure}[t]
\centering
\includegraphics[width=3.4in]{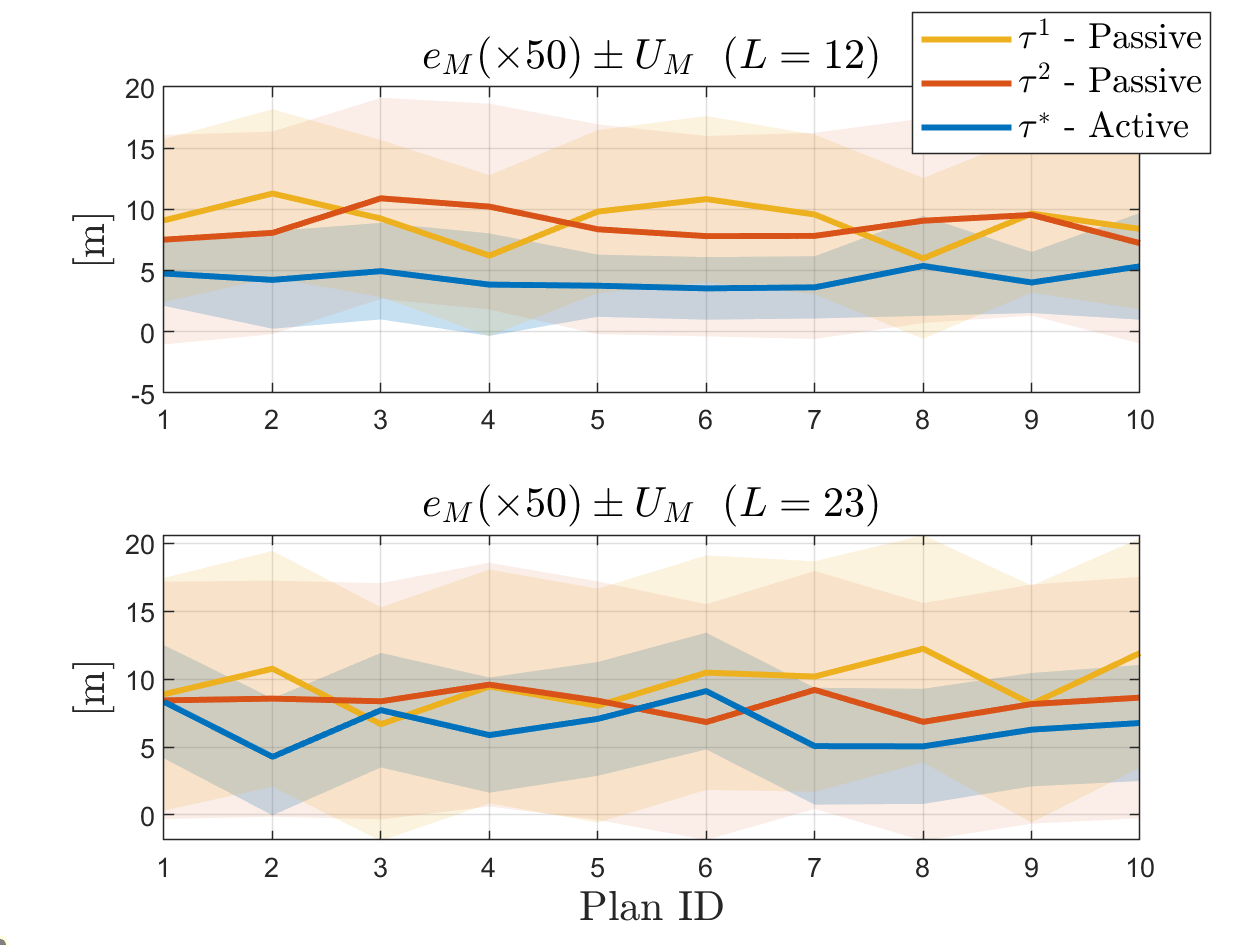}
\caption{Mean map estimation performance on time horizons $L_1$ and $L_2$. 
Solid lines represent the landmark estimation error, shaded areas represent the uncertainty.}
\label{fig:map_est_perf}
\end{figure}
\begin{figure}[t]
\centering
\includegraphics[width=3.4in]{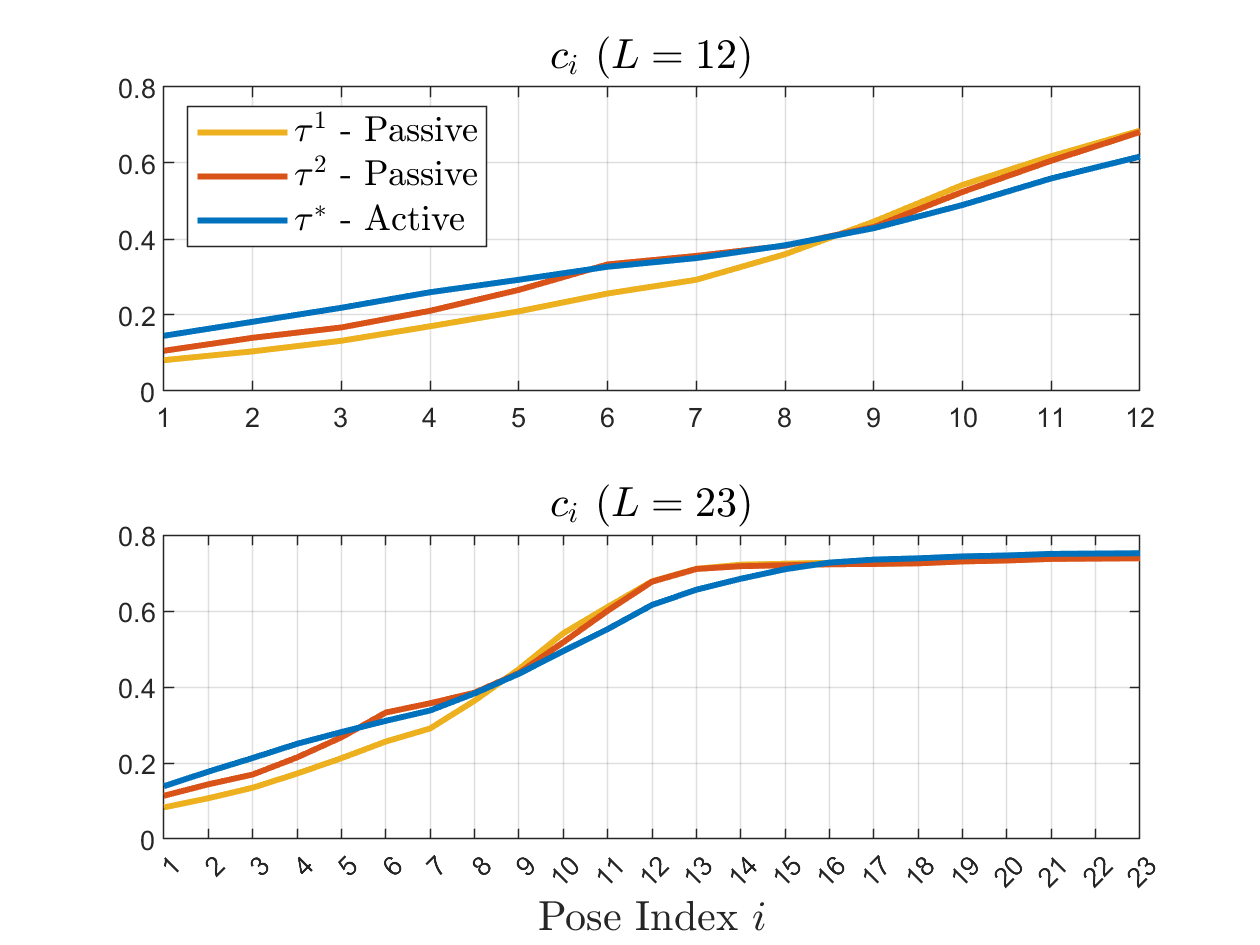}
\caption{Mean map coverage on time horizons $L_1$ and $L_2$.}
\label{fig:map_cover}
\end{figure}
The proposed vision-based active SLAM approach is evaluated by simulating spacecraft proximity SLAM and comparing the results obtained along three different families of trajectories, $\tau^1=T_{k+1:k+L}^1, \tau^2=T_{k+1:k+L}^2$ and $\tau^*=T_{k+1:k+L}^*$.
Trajectories $\tau^1$ and $\tau^2$ are derived from fixed passive observation goals $\mathbf{r}_O^1=[0, 0, 2]^{\top}$ and $\mathbf{r}_O^2=[0,0, 0]^{\top}$ corresponding to the geometrical center of the HST model and to the origin of the target frame $\mathcal{T}$, respectively.
On the other hand, $\tau^*$ performs pointing of an informative target $\mathbf{r}_O^*$ given by the active SLAM pipeline.
The SLAM results $\mathcal{X}^{\textrm{MAP},1}$, $\mathcal{X}^{\textrm{MAP},2}$ and $\mathcal{X}^{\textrm{MAP},*}$ obtained in the three cases above (see examples in Figs.~\ref{fig:tau1_3D}, \ref{fig:tau2_3D} and \ref{fig:tau3_3D}) are compared in terms of estimation uncertainty and accuracy.
We also compute the mean map coverage in the different cases as it provides further insights on the results.
While our method only tackles uncertainty reduction in an explicit way, we show that it also yields higher accuracy when compared to passive pointing strategies.

To account for possible performance fluctuations due to random effects introduced by process and model noise, we evaluate ten instances of our method, hence computing ten different informative plans and comparing each one of them with both $\tau^1$ and $\tau^2$.
Moreover, for each of the ten comparisons above we simulate SLAM ten times on the time window $t_{k+1:k+L}$.
In this way, it is possible to assess the efficacy of our active SLAM method by examining the average performance of each plan.
The above analysis is reported for two different planning horizons $L_1$ and $L_2$ (see Table \ref{tab:activeSLAM_params}), corresponding to $1/5$ and $2/5$ of the chaser orbit, respectively.

Several performance indices are established for the pose and map variables to assess performance across the different methods.
Given a simulated SLAM instance identified by index $q$, $q=1,\ldots,10$, the following quantities are defined.
\paragraph{Estimation Uncertainty}
\begin{equation}
\label{eq:traj_cov_trace}
    U_{\mathbf{r}_i}^q = \mathrm{Tr}(\Sigma_{\mathbf{r}_i}^q), \quad
    U_{\bm{\phi}_i}^q = \mathrm{Tr}(\Sigma_{\bm{\phi}_i}^q), \quad
    U_{M}^q = \frac{1}{N_l} \sum_{j=1}^{N_l} \mathrm{Tr}(\Sigma_{\mathbf{l}_j}^q),
\end{equation}
where $N_l$ is the number of triangulated landmarks, and $\Sigma_{\mathbf{r}_i}^q$ ($\Sigma_{\bm{\phi}_i}^q$) and $\Sigma_{\mathbf{l}_j}^q$ are the marginal covariance matrices associated to the estimates of the $i$-th position (attitude) and $j$-th landmark, respectively, during simulation $q$.
\paragraph{Estimation Accuracy}
\begin{equation}
\label{eq:pos_err_rel}
    \mathrm{e}_{\mathbf{r}_i}^q =  \lVert \mathbf{r}_i^q - \hat{\mathbf{r}}_i^q\rVert, \quad
    \mathrm{e}_{\bm{\phi}_i}^q =  \lVert \bm{\phi}_i^q - \hat{\bm{\phi}}_i^q\rVert, \quad
    \mathrm{e}_{M}^q = \frac{1}{N_l} \sum_{j=1}^{N_l}  \lVert \mathbf{l}_j^q - \hat{\mathbf{l}}_j^q\rVert,
\end{equation}
where $\mathbf{r}_i^q$ ($\bm{\phi}_i^q$) and $\hat{\mathbf{r}}_i^q$ ($\hat{\bm{\phi}}_i^q$) are the ground truth and estimated values of the $i$-th camera position (attitude), respectively, and similarly for the true and estimated $j$-th landmark position, $\mathbf{l}_j^q$ and $\hat{\mathbf{l}}_j^q$.
The attitude is described by the yaw-pitch-roll sequence $\bm{\phi}_i \in \mathbb{R}^3$.
\paragraph{Map Coverage}
The map coverage index $c_i^q$ is defined as a function of time step $i$ along the $q$-th simulated trajectory,
\begin{equation}
    \label{eq:map_coverage}
    c_i^q = \frac{\mathrm{dim}(M_i^q)}{\mathrm{dim}(M^q)}.
\end{equation}
In Eq.~\eqref{eq:map_coverage}, $M_i^q$ collects the map landmarks that have been seen by the camera up to the $i$-th time step. 
Variable $M^q$ collects all the landmarks observed during the reconnaissance phase prior to simulation $q$, and acts as an heuristic upper bound on the number of landmarks that can be triangulated.

The mean values of $U_{\mathbf{r}_i}^q$, $U_{\bm{\phi}_i}^q$, $e_{\mathbf{r}_i}^q$, $e_{\bm{\phi}_i}^q$ and $c_q^i$ are then computed over $10$ different simulations for each of the $10$ plans while keeping the value of $i$ (the pose/time index) fixed, to visualize the time evolution for each of the pointing strategies, as follows,
\begin{align}
    \label{eq:mean_pos_error}
U_{\mathbf{r}_i} &= \frac{1}{100} \sum_{\textrm{Plan ID = 1}}^{10} \sum_{q=1}^{10} U_{\mathbf{r}_i}^q, \quad
e_{\mathbf{r}_i} = \frac{1}{100} \sum_{\textrm{Plan ID = 1}}^{10} \sum_{q=1}^{10} e_{\mathbf{r}_i}^q, \\
    \label{eq:mean_att_error}
    U_{\bm{\phi}_i} &= \frac{1}{100} \sum_{\textrm{Plan ID = 1}}^{10} \sum_{q=1}^{10} U_{\bm{\phi}_i}^q, \quad
    e_{\bm{\phi}_i} = \frac{1}{100} \sum_{\textrm{Plan ID = 1}}^{10} \sum_{q=1}^{10} e_{\bm{\phi}_i}^q,
\end{align}
\begin{equation}
\label{eq:mean_cover}
    c_i = \frac{1}{100} \sum_{\textrm{Plan ID = 1}}^{10} \sum_{q=1}^{10} c_i^q.
\end{equation}
The map estimation performance is, instead, averaged across the different simulations to compute its mean for each plan,
\begin{equation}  
\label{eq:map_mean_perf}
    U_{M} = \sum_{q=1}^{10} U_M^q, \quad
    e_M = \sum_{q=1}^{10} e_M^q.
\end{equation}
The values found using Eqs.~\eqref{eq:mean_pos_error}-\eqref{eq:mean_att_error} are reported in Fig.~\ref{fig:pose_performance}.
The mean pose uncertainty, expressed as the trace of the covariance matrix, is always smaller for our active strategy (shaded blue region) than it is for the passive ones (yellow and red regions), both for position and attitude variables.
Specifically, the uncertainty in the proposed active SLAM framework is $\sim 50\%-70\%$ smaller than the one achieved in the simulated passive scenarios.
As expected, the attitude uncertainty is constant since the orientation of the chaser is controlled, while the trace of the position covariance grows over time given that no position control is applied and loop closure cannot be performed.
Moreover, our proposed approach also leads to better accuracy (smaller average error), as shown by the blue lines in Figs.~\ref{fig:pos_perf_L1}-\ref{fig:pos_perf_L2}.

Similar results are obtained for the map estimation accuracy and uncertainty, displayed in Fig.~\ref{fig:map_est_perf}.
While better confidence and accuracy are always achieved for the shortest time horizon $L_1$, the average landmark estimation error oscillates more when $L=L_2$, and in two cases (plan IDs $3$ and $6$) it is higher than the one resulting from one of the two passive strategy.
Such behavior is to be expected as the process noise affects more the planning performance when the planning horizon increases.
A 3D representation of the map estimation result with the correspondent uncertainties for trajectories $\tau^1$, $\tau^2$ and $\tau^*$ is also provided in Fig.~\ref{fig:SLAM_result}.

For the planning interval $L_1$, the active strategy ``explores'' less than the other two, covering a slightly smaller fraction of the total map (Fig.~\ref{fig:map_cover}).
This can be explained since, for a shorter lookahead window, inserting more landmarks in the estimation graph without having the chance to accumulate multiple observations would lead to an increase in the overall uncertainty.
On the other hand, a different behavior is observed for the longer planning window $L_2$, where the planner covers a portion of the scene comparable, if not superior, to the one resulting from the passive cases.
Indeed, by inspecting Fig.~\ref{fig:map_cover} for $L=23$, it can be seen that the planner explores after the chaser has reached the first half of its path, after $i=13$.
Then, the flatter region after $i=16$ indicates that fewer new landmarks are being observed towards the end of the plan, allowing the system to accurately estimate the position of those already present in the SLAM graph.

\section{Conclusions}
In this paper, we introduced an information-theoretic approach to active visual SLAM, specifically tailored for spacecraft proximity operations. 
Our method is grounded in the use of factor graphs as the probabilistic framework for the SLAM problem, and it utilizes the \texttt{GTSAM} library for graph optimization. 
The proposed approach marks a departure from previous works by establishing a rigorous link between the pre-existing knowledge of the environment, the spacecraft-mounted camera's future trajectory, and the resulting quality of the SLAM estimate.
We evaluated candidate attitude trajectories based on their potential to reduce joint pose and map uncertainty, quantified as the difference in entropy before and after the trajectory execution. 
Our simulation results demonstrate that the active attitude trajectories significantly enhance the confidence and accuracy of the SLAM solution compared to passive approaches.
Some promising avenues for further research include extending our current attitude-only strategy to encompass camera position optimization and integrating our approach with advanced visual SLAM pipelines, possibly including different types of sensors such as gyroscopes and star trackers.

\section{Acknowledgment}
The authors would like to acknowledge the generous support of NSF (FRR award no. 2101250) and AFOSR (SURI award no. FA9550-23-1-0723).

\bibliographystyle{AAS_publication}   
\bibliography{references}

\end{document}